\pdfoutput=1

\documentclass[11pt]{article}

\usepackage[final]{acl}

\usepackage{times}
\usepackage{latexsym}
\usepackage{amssymb}
\usepackage{amsmath}

\usepackage{booktabs}
\usepackage{multirow}
\usepackage{graphicx}

\usepackage{caption}
\usepackage{subcaption}
\usepackage{enumitem}
\usepackage{tabularx}

\usepackage{xcolor}
\usepackage{colortbl}
\usepackage{booktabs}
\usepackage{array}

\definecolor{junkcolor}{RGB}{225,225,225}
\definecolor{genericcolor}{RGB}{204,229,255}
\definecolor{proximatecolor}{RGB}{255,242,204}
\definecolor{targetcolor}{RGB}{212,239,223}
\definecolor{answercolor}{RGB}{169,209,142}

\usepackage[T1]{fontenc}

\usepackage{microtype}

\usepackage{inconsolata}
\usepackage[most]{tcolorbox}

\newtcolorbox{sharp_box}{
    sharpish corners, 
    boxrule = 0pt,
    toprule = 4.5pt, 
    enhanced,
    fuzzy shadow = {0pt}{-2pt}{-0.5pt}{0.5pt}{black!35} 
}

\title{CulTrace: Tracing Internal Cultural Reasoning in Large Language Models}

\def\authorsep{\hspace{0.3em}}

\author{Haeun Yu\textsuperscript{1} \authorsep Arnav Arora\textsuperscript{1} \authorsep Seogyeong Jeong\textsuperscript{2} \authorsep Nadav Borenstein\textsuperscript{1} \\ \textbf{Siddhesh Pawar\textsuperscript{1}} \authorsep \textbf{Jisu Shin\textsuperscript{2}} \authorsep \textbf{Jiho Jin\textsuperscript{2}}
\textbf{Junho Myung\textsuperscript{2}} \\ \textbf{Alice Oh\textsuperscript{2}} \authorsep \textbf{Isabelle Augenstein\textsuperscript{1}} \medskip\\
\null\textsuperscript{1}University of Copenhagen \quad \null\textsuperscript{2}KAIST\\
\texttt{hayu@di.ku.dk}}

\begin{document}
\maketitle
\begin{abstract}

The growing deployment of large language models (LLMs) across diverse cultural contexts necessitates a deeper understanding of models' hidden representations of different cultures. Prior work has evaluated cultural awareness in LLMs by analysing their outputs. This approach overlooks how cultures are represented within the model parameters, missing why models generate incorrect responses.  To bridge this gap, we propose \textbf{CulTrace}, a mechanistic interpretability-based method that probes the internal representations of LLMs for cultural knowledge. With CulTrace, we inspect how cultural knowledge is processed across layers and how it is integrated during cultural QA. We find a consistent staged trajectory of cultural reasoning. Models first engage with the question's domain, then resolve the relevant culture, and finally narrow in on an answer. We also demonstrate that models' cultural reasoning is imbalanced, showing delayed relevant culture resolution and more confusion with less-represented cultures.

\end{abstract}

\section{Introduction}

\begin{figure}[t]
    \centering
    \includegraphics[width=\columnwidth]{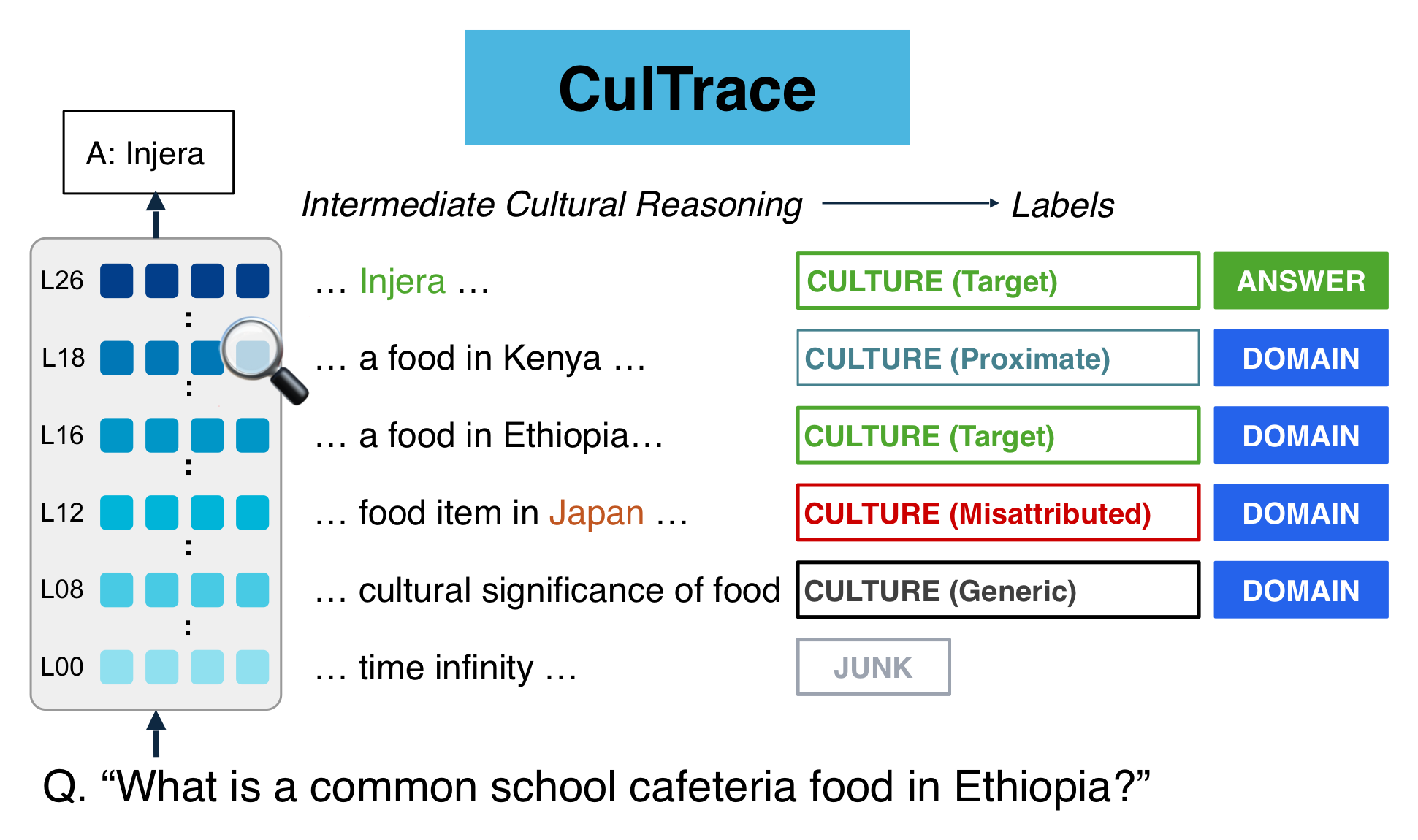}
    \caption{We present an overview of the proposed framework. Given a cultural question about a target culture (Ethiopia), CulTrace decodes the model's internal representations at each layer to reveal what the model encodes, spanning from topic (e.g., food) and answer candidates (e.g., injera) to cultural context (e.g., misattributed cultures such as Kenya or Japan).}
    \label{fig:overview}
\end{figure}

Cultural alignment has become an important aspect of LLM development due to their worldwide deployment. Most existing evaluations of cultural alignment are extrinsic, measuring model outputs on cultural understanding or simulation tasks. While these approaches can reveal \textit{when} models produce culturally inaccurate responses, they provide limited insight into \textit{why}, making it challenging to improve performance. To address this gap, we introduce \textbf{CulTrace}, a diagnostic mechanistic interpretability framework that verbalises layer-wise model representations of cultural reasoning into natural language. Using CulTrace, we output chain of reasonings across layers within a model, allowing us to inspect when the model is focusing on various elements pertaining to a query, e.g., the domain of the question, cultural context, candidate answers and more (Figure~\ref{fig:overview}).

CulTrace (\S\ref{sec:culturescope}) builds on LatentQA \cite{pan2026latentqa}, a method for decoding model activations into natural language by training decoders that take model activations and a question as input and a corresponding answer as output. The trained decoder generalises to tasks beyond training, allowing one to successfully probe the activations for various tasks like recovering hidden system prompts, predicting future generations, or extracting relational knowledge through a decoding question. While \citet{pan2026latentqa} use a specific layer from the model to decode activations, we adapt it by training decoders for each layer and creating a chain of intermediate reasoning across the model's layers. This helps verbalise how information emerges and is refined across the layers (Table \ref{tab:example}). We apply this framework to cultural QA as it is a challenging task with subjective answers compared to domains like math or code, making it crucial to understand intermediate reasoning and cultural bias. 

We employ CulTrace for two cultural probing tasks, tracing intermediate cultural reasoning and identifying the target culture, when answering a cultural question. For the former (\S\ref{sec:cul_reasoning}), we generate intermediate reasoning at each layer and characterise whether the model is processing the question's domain, cultural context, or answer selection, amongst others. For the latter (\S\ref{sec:imbalance}), we probe for which culture the model is encoding at each layer while processing a culturally grounded question, allowing us to compare model behaviour across cultures and analyse cultural confusions. We show the effectiveness of our method across three widely used open-weight LLMs, using two culturally grounded question-answering datasets, BLEnD \cite{myung2024blend} and CANDLE QA \cite{candle2023}, both covering ten cultures and various facets of culture.

Our results reveal a consistent staged trajectory of cultural reasoning. Models first engage with the question's domain, then resolve the relevant culture, and finally narrow in on an answer. While domain engagement begins at similar depths across cultures, the layer at which the target culture is resolved varies, with less-represented cultures tending to be resolved later. 
Tracing this broad-to-specific process also reveals how cultural confusions emerge internally. Before resolving the target culture, models often represent a broader geographical region, a proximate culture, or a model-specific default culture. Such intermediate associations occur more frequently for less-represented cultures, indicating that cultural disparities are reflected not only in model outputs but also in how cultural knowledge is retrieved and resolved across internal layers. Our findings suggest that cultural alignment interventions could target the layers at which cultural context is being resolved, rather than operating solely on final outputs. In sum, our contributions are:

\begin{enumerate}[noitemsep]
    \item We present the first diagnostic framework using mechanistic interpretability to identify and study patterns of cultural reasoning across layers in the model.
    \item We train and release layer-specific decoders for three models, allowing us to probe various cultural and non-cultural knowledge across the layers in those LLMs.
    \item We create a dataset and taxonomy of intermediate cultural reasoning, tracing where across the model layers reasoning related to various parts of the query happens\footnote{Codes, models, datasets are available at \url{https://github.com/copenlu/CulTrace}}.
    \item Our analysis produces insights into intermediate cultural reasoning and bias amongst cultures within the layers of the model, supporting evidence from existing studies conducting extrinsic analysis.
\end{enumerate}

\section{Related Work}

\paragraph{Cultural Understanding in LLMs}

Previous work has proposed evaluation datasets and frameworks to assess LLMs' cultural understanding ability acquired during pre-training \cite{keleg-magdy-2023-dlama, naous-xu-2025-origin, pawar2025survey}. BLEnD \cite{myung2024blend} provides a multilingual and multi-domain commonsense QA dataset spanning 16 cultures and regions, with six topics, designed to uncover cross-cultural disparities in everyday knowledge. Other multilingual benchmarks \cite{zhou-etal-2025-mapo, hasan-etal-2025-nativqa, wang-etal-2024-seaeval, cao-etal-2024-cultural} construct culturally localised evaluation datasets that span domains such as cuisine, proverbs, news, and reasoning. Across these datasets, performance gaps are consistently observed between high-resource and underrepresented languages and cultures, often linked to pre-training data imbalances that favour dominant regions \cite{naous-xu-2025-origin}. While these efforts highlight important cross-cultural disparities, they perform an extrinsic evaluation, overlooking the underlying mechanism and cultural knowledge space embedded in LLMs. To address this gap, our paper aims to reveal how cultural knowledge is embedded in models' internal representations.

\paragraph{Mechanistic Interpretability}

MI seeks to explain LLM behaviour in terms of internal structure, from individual components such as neurons and attention heads~\cite{meng2023locating, geva-etal-2023-dissecting, yu-etal-2024-revealing} to natural-language verbalizers that render representations as text~\cite {hernandez2024inspecting, geva-etal-2022-lm, belrose2025tunedlens}. Early verbalizers such as the logit lens and Patchscope are restricted to single-token projections and cannot capture information distributed across multiple tokens~\cite{patchscope, belrose2025tunedlens}. LatentQA~\cite{pan2026latentqa} instead trains a decoder to answer questions directly over model activations spanning multiple tokens, providing a more flexible probe that leverages the language capabilities of LLMs.

Recent work applies MI to culture: isolating culture-specific neurons~\cite{namazifard-poech-2025-isolating}, extracting cultural steering vectors~\cite{veselovsky-etal-2026-localized}, and discovering interpretable cultural features with sparse autoencoders~\cite{khanuja2026steeringllmsculturallylocalized}. These methods identify where cultural information is stored or how to steer it, but they treat cultural processing within models as a static property of components or directions. Going beyond the localisation of cultural knowledge, our work traces the cultural reasoning throughout the forward pass, characterising each layer's role in LLMs' cultural reasoning.

\begin{figure}[t]
    \centering
    \includegraphics[width=1\columnwidth]{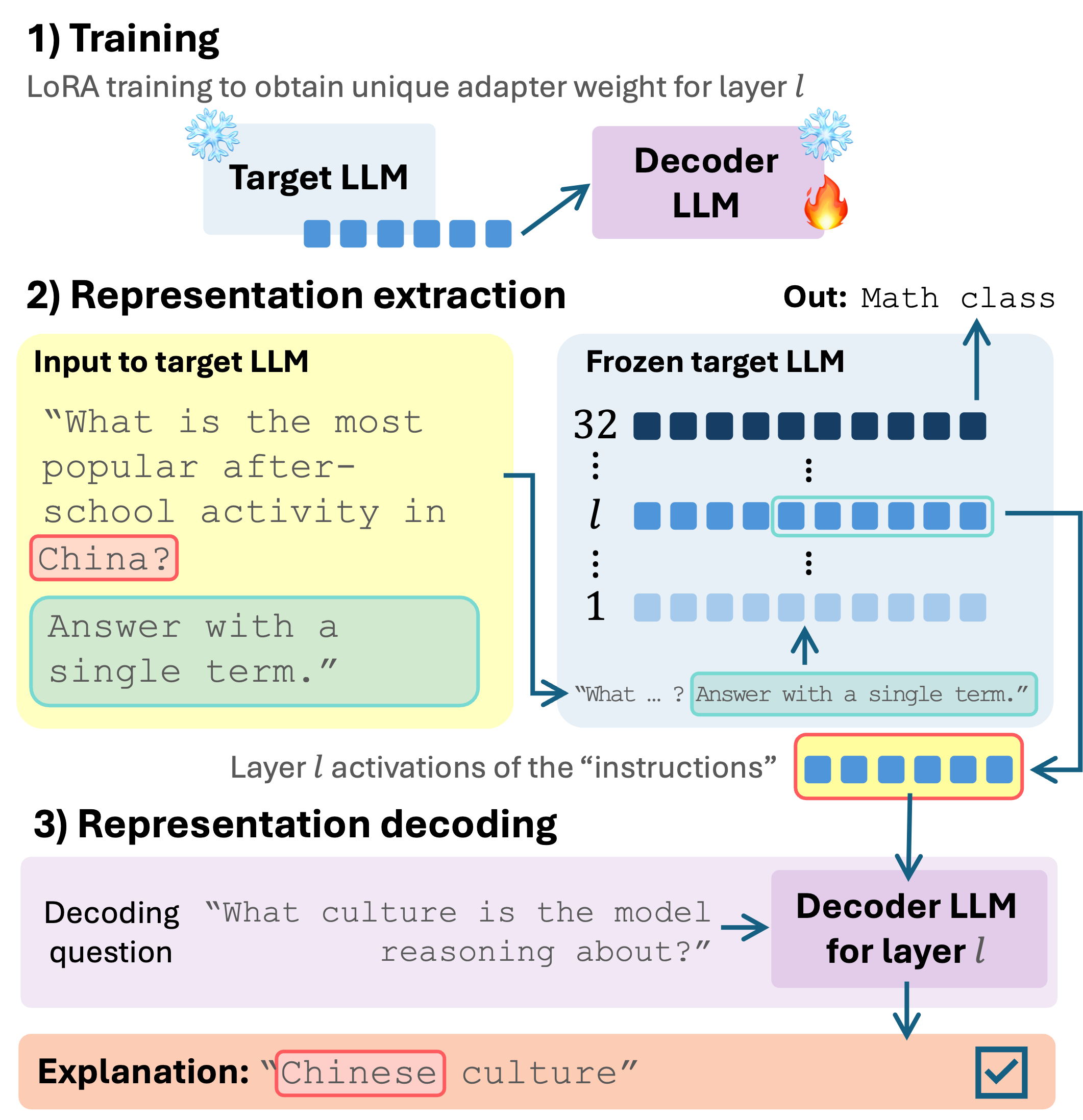}
    \caption{CulTrace first trains the decoder LLM for a unique adapter weight with a shared base weight by LoRA at the \textbf{Training} stage. After training, in the \textbf{Representation extraction} stage, the target LLM generates an answer to the question, and we extract the representation at layer $l$. Then, during \textbf{Representation decoding}, with a decoding question, we translate the representation into natural language. Here, we show example output where the decoder model is asked to identify the target culture (China).}
    \label{fig:culturescope}
\end{figure}

\section{CulTrace: Tracing Cultural Reasoning within Internal Layers}
\label{sec:culturescope}

Translating LLMs' internal representations into natural language can reveal knowledge stored in the cultural concept space. We devise CulTrace, building on LatentQA \cite{pan2026latentqa}, an established method that trains a decoder model to answer questions directly over another model's activations. 

LatentQA involves two models: a target model $\mathcal{M}$, whose internal representations are to be explained, and a decoder model $\mathcal{D}$, trained on paired activations and QA examples to translate those representations into natural language. At inference time, $\mathcal{M}$ processes an input and produces a hidden representation; this representation is patched into $\mathcal{D}$, and a \textit{decoding question} guides $\mathcal{D}$ to describe a specific aspect of the representation in natural language. For example, a decoding question can be about predicting what persona the target model is going to generate \citet{pan2026latentqa}.
CulTrace consists of three stages that adapt LatentQA to enable layer-wise cultural reasoning tracing.

\subsection{Method}
\label{subsec:method}

\paragraph{Step 1. Training}

In LatentQA, the layer from which representations are extracted is treated as a hyperparameter and is fixed during evaluation. Since we are interested in tracing cultural reasoning across all layers, we train separate decoders for each layer of the target model. We always patch into layer $0$ of the decoder. To keep this tractable, trained decoder models share the base decoder parameters, and we train only a lightweight LoRA adapter per layer. At inference time, decoding from a different layer requires only swapping the adapter. Following LatentQA, we validate the trained decoders on recovery of factual associations (Appendix \ref{appx:loss}). For more technical details and training details, we refer to Appendix \ref{appx:training_detail}.

\paragraph{Step 2. Representation Extraction}

The target model $\mathcal{M}$ encodes a cultural question $q$ about a target culture $c$ and an instruction $\mathcal{I}$ (``Answer with a single term.'' in Figure \ref{fig:culturescope}) and generates an answer followed by its rationale. Generating a rationale encourages representations that encode the cultural knowledge supporting the answer, not only the answer itself. We then extract the hidden state of $\mathcal{I}$, denoted $x_l^\mathcal{I}$, rather than the full sequence: this prevents the decoder from reading $q$ directly and ensures it relies on the $\mathcal{M}$'s internal state. We show the impact of extraction position in Appendix \ref{eval:pos}. Extracting $x_l^\mathcal{I}$ at every layer lets us trace how culturally relevant knowledge is retrieved and refined through the forward pass. The input prompt is provided in Figure \ref{fig:appendix_Prompts}.

\paragraph{Step 3. Representation Decoding}

At this stage, the decoder model generates an answer (Chinese culture in Figure \ref{fig:culturescope}) to a decoding question (``What culture is the model reasoning about?'' in Figure \ref{fig:culturescope}) using the patched representation $x_l^\mathcal{I}$. We design decoding questions that reveal how cultural knowledge is retrieved and refined across layers. Following LatentQA, the prompt starts with a sequence of placeholder tokens `?' of a length matching the internal representations to be patched. At layer $0$, we replace the hidden representation of the placeholder tokens with $\boldsymbol{x^\mathcal{I}}$.

With CulTrace, we can generate chain of reasonings across layers within a model. This allows us to inspect how the model processes its parametric knowledge while answering culturally grounded questions. We examine models' internal representations of cultural knowledge using two cultural QA datasets. 

\subsection{Cultural QA Datasets}
\label{section:datasets}

We select BLEnD \cite{myung2024blend}, a cultural commonsense QA dataset. BLEnD is a hand-crafted multilingual benchmark designed to evaluate LLMs' everyday knowledge across diverse cultures. In addition, we use CANDLE, which is a comprehensive cultural commonsense knowledge base that provides 1.1m assertions about various cultures sourced from CommonCrawl \cite{candle2023}. Specifically, we use the filtered version of CANDLE from \citet{pawar-etal-2025-presumed}, which retains only assertions with specific, distinctive cultural knowledge. We convert those assertions into QA format using an LLM (Qwen3.6-27B-FP8 \cite{qwen3.6-27b}). An example assertion is ``\textit{Sangria is a Spanish wine cocktail made of red or white wine and fresh fruit}''. The resulting converted question is ``\textit{What is the Spanish wine cocktail made of red or white wine and fresh fruit?}''. The input prompt and validation details for conversion are in Appendix \ref{appx:cloze_to_qa}. We refer to the created dataset as CANDLE QA. We use English questions for both datasets as LatentQA is trained for English. To keep a similar culture distribution between the two datasets, we select ten cultures to study (Table \ref{tab:datasets}). Dataset details can be found in Appendix~\ref{appendix:dataset}.

\subsection{Models}

As MI methods require access to full model parameters, we use three popular open‑weight LLMs: Meta-Llama-3.1-8B-Instruct (\texttt{Llama-8b}, \citet{grattafiori2024llama3herdmodels}), gemma-3-4b-it (\texttt{gemma-4b}, \citet{gemmateam2025}), and 
Ministral-8B-Instruct-2410 (\texttt{Ministral-8b}\footnote{\url{https://huggingface.co/mistralai/Ministral-8B-Instruct-2410}}).

\begin{table*}[t!]
\centering
\scriptsize
\begin{tabular}{@{}p{0.5cm}|p{10.5cm}|p{3.5cm}@{}}
\toprule
\multicolumn{3}{@{}p{15.5cm}@{}}{\textbf{Question:} What are the most commonly eaten snacks at shopping malls in South Korea?} \\[3pt]
\multicolumn{3}{@{}p{14.5cm}@{}}{\textbf{LLM Answer:} Hotteok — Reasoning: Hotteok is a popular Korean-style pancake-like snack that is commonly sold at shopping malls in South Korea. It is typically filled with sweet fillings such as honey, sugar, and cinnamon, or savory fillings.} \\
\midrule
\textbf{Layer} & \textbf{Reasoning Trace} & \textbf{LLM labels} \\
\midrule
\rowcolor{junkcolor}
0  & The assistant is contemplating the concept of infinity, focusing on its abstract and paradoxical nature. & \textit{Junk} \\
\rowcolor{junkcolor}
2  & The assistant is contemplating the concept of time and its perception. & \textit{Junk} \\
\rowcolor{junkcolor}
4  & The assistant is contemplating the concept of time and its subjective nature. & \textit{Junk} \\
\rowcolor{genericcolor}
6  & The assistant is contemplating the concept of a \textbf{cultural phenomenon} that is both widely accepted and criticized. & \textit{Culture (Unspecified)} \\
\rowcolor{genericcolor}
8  & The assistant is contemplating the \textbf{cultural significance} of a particular \textbf{snack} and its widespread appeal. & \textit{Culture (Unspecified), Domain} \\
\rowcolor{proximatecolor}
10 & The assistant is contemplating the \textbf{cultural significance} of a particular \textbf{snack} in \textbf{Japan}. & \textit{Culture (Proximate), Domain} \\
\rowcolor{proximatecolor}
12 & The assistant is contemplating the most \textbf{popular snack} among \textbf{Japanese} people. & \textit{Culture (Proximate), Domain} \\
\rowcolor{proximatecolor}
14 & The assistant is contemplating the most \textbf{popular snack} among \textbf{students} in \textbf{Japan}. & \textit{Culture (Proximate), Domain} \\
\rowcolor{genericcolor}
16 & The assistant is contemplating the reasons why \textbf{snacks are popular} among students during exams. & \textit{Domain} \\
\rowcolor{targetcolor}
18 & The assistant is contemplating the cultural significance of \textbf{Korean snacks}, particularly the popular \textbf{Korean Fried Chicken}. & \textit{Culture (Target), Domain, Answer (Candidate)} \\
\rowcolor{targetcolor}
20 & The assistant is contemplating the cultural significance of \textbf{Korean snacks} and how they have become a popular aspect of \textbf{Korean culture}. & \textit{Culture (Target), Domain} \\
\rowcolor{answercolor}
22 & \textbf{The assistant is contemplating the popular Korean snack, \textit{Hotteok}, and its widespread appeal.} & \textbf{\textit{Culture (Target), Domain, Answer (Generated)}} \\
\rowcolor{targetcolor}
24 & The assistant is contemplating the reasons why \textbf{Korean-style convenience stores} have become so popular globally. & \textit{Culture (Target), Domain} \\
\rowcolor{targetcolor}
26 & The assistant is contemplating the cultural significance of \textbf{snacks in Korean society} and how they contribute to social gatherings. & \textit{Culture (Target), Domain} \\
\rowcolor{targetcolor}
28 & The assistant is contemplating the \textbf{cultural significance of snacks in Korean society} and how they contribute to social interactions. & \textit{Culture (Target), Domain} \\
\rowcolor{genericcolor}
30 & The assistant is contemplating the reasons behind the \textbf{user's preference for certain foods}, seeking to understand the underlying motivations. & \textit{Domain} \\
\bottomrule
\end{tabular}

\caption{Example outputs using CulTrace for even-number layers. \textbf{Bold} text in the Reasoning Trace column denotes key phrases driving the label; \textit{italic} text in the LLM judge labels column gives the annotated category. Row color indicates label category: \colorbox{junkcolor}{\textit{Junk}} (off-topic reasoning), \colorbox{genericcolor}{\textit{Culture (Unspecified)}} (culture-relevant but unanchored), \colorbox{proximatecolor}{\textit{Culture (Proximate)}} (anchored to a related-but-incorrect culture), \colorbox{targetcolor}{\textit{Culture (Target)}} (anchored to the correct culture, South Korea), and \colorbox{answercolor}{\textbf{Answer (Generated)}} (final answer emerges). \textit{Domain} indicates the trace references the snack/question topic and \textit{Answer (Candidate)} indicates a candidate answer is being considered.}
\label{tab:example}
\end{table*}

\begin{figure}
    \centering
    \includegraphics[width=1\columnwidth]{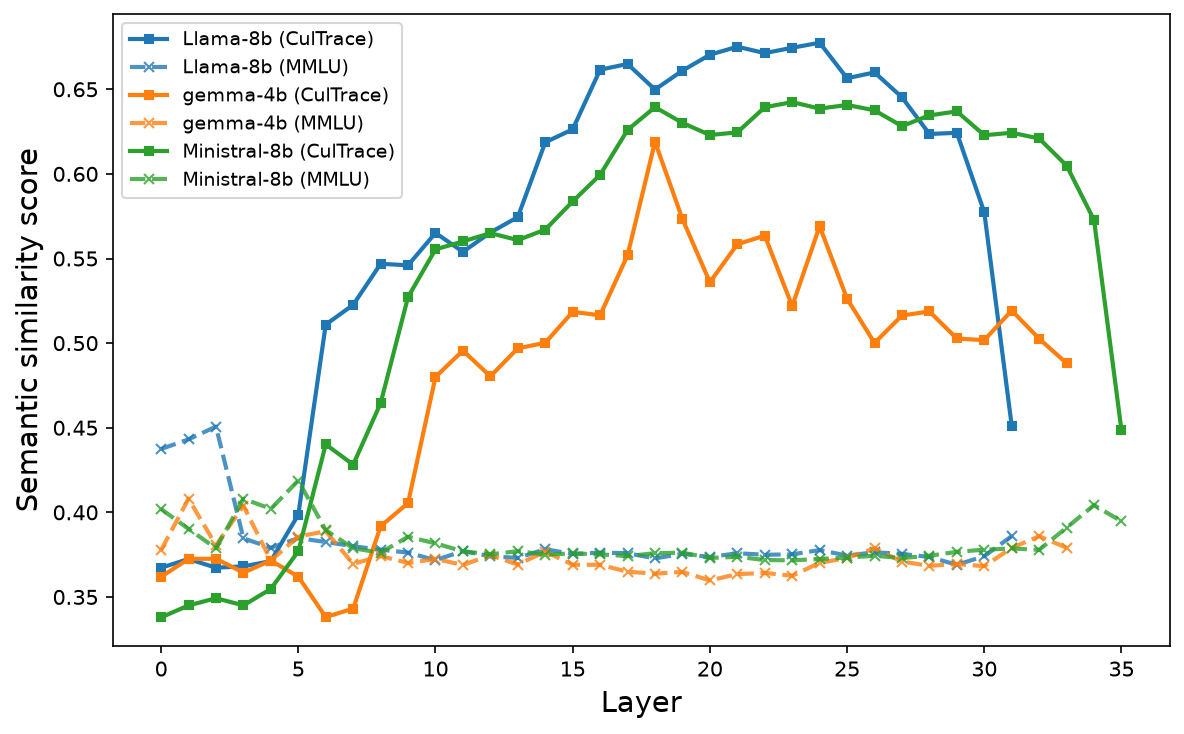}
    \caption{We compare the semantic similarity of CulTrace across layers with different set-ups. The result shows that CulTrace decodes cultural reasoning specific to cultural questions (Figure \ref{fig:recovery_mmlu}), demonstrating the method's domain transferability.}
    \label{fig:recovery_mmlu}
\end{figure}

\subsection{Evaluation of CulTrace}
\label{sec:eval_culturescope}

LatentQA demonstrates its effectiveness for verbalising future predictions from hidden representations, recovering hidden system prompt personas from user prompt representations with over 80\% accuracy. Since it has not been evaluated in the culture domain, we test whether this verbalisation competence can transfer to the culture domain to validate CulTrace. We assess how faithfully CulTrace recovers the model's reasoning for its answer from internal representations. Just as the hidden persona serves as ground truth in the persona task, the model's generated reasoning serves as ground truth here. If CulTrace can recover reasoning from an intermediate layer representation extracted before the reasoning is generated, then the representation genuinely encodes the reasoning path the model follows, and the decoder model is capable of capturing it.

We set the decoding question to ``What is the answer and why?'' and compute the semantic similarity between the reasoning decoded by CulTrace and the reasoning generated by the target LLM as part of the output. To calculate semantic similarity, we first encode the reasoning into embeddings using an embedding model.\footnote{\url{https://huggingface.co/google/embeddinggemma-300m}} Then, we measure cosine similarity between the two sequence embeddings. 

As ablations, we compare the semantic similarity of the generated reasonings from CulTrace with those generated by patching representations from MMLU science questions \cite{hendryckstest2021} instead of cultural questions\footnote{We use questions from college physics, biology, and chemistry subtopics.}. If the decoder model were generating from the decoding question alone rather than the patched representation, these out-of-domain representations would match the cultural explanation as well as CulTrace does. We expect higher semantic similarity scores for CulTrace than for both random and out-of-domain representations.

Figure \ref{fig:recovery_mmlu} shows per-layer semantic similarity scores compared with those from MMLU science questions. We present results with Gaussian noise vectors in Appendix \ref{appendix:noise}. \textit{CulTrace achieves higher semantic similarity scores than random and out-of-domain representations from layer 6 onwards with three different models.} We also present outputs from ablations in Table \ref{tab:mmlu_gemma_example_a} and Table \ref{tab:mmlu_gemma_example_b}.

\section{Intermediate Cultural Reasoning}
\label{sec:cul_reasoning}

In Section \ref{sec:eval_culturescope}, we show that CulTrace can faithfully extract the future generations from its intermediate representations, prior to generating them. With the validity of the representation verbalising method established, we now use CulTrace to trace how the model retrieves and refines cultural knowledge across layers by verbalising intermediate reasoning in each layer.

\begin{table}[t]
\centering
\small
\begin{tabular}{p{1.1cm}p{1.4cm}p{4.2cm}}
\toprule
\textbf{Dim.}       & \textbf{Label}    & \textbf{Description}                                        \\ \midrule
\multirow{4}{*}{Culture} & Unspecified       & Mentions culture without specific culture name \\
                         & Misattributed     & Mentions a culture unrelated to the target culture          \\
                         & Proximate         & Mentions a culture similar to the target culture            \\
                         & Target            & Mentions a culture the question is about                    \\
\midrule
Domain                    & $\checkmark/~\times$ & Engages with the question's topic area                      \\
\midrule
\multirow{2}{*}{Answer}  & Generated         & Mentions the model's own answer                             \\
                         & Candidate         & Mentions a plausible alternative      \\
\midrule
Reflection                & $\checkmark/~\times$ & Meta-cognitive evaluation of the model's own reasoning      \\ \midrule
Syntax                   & $\checkmark/~\times$ & Processing of task format or output planning                \\ \midrule
Junk                     & $\checkmark/~\times$ & Unrelated to the given task                                 \\ \bottomrule
\end{tabular}%
\caption{Dimensions, labels, and descriptions for intermediate cultural reasoning characterisation across layers. $\checkmark/~\times$ represent binary labels of presence.}
\label{tab:taxonomy}
\end{table}

\subsection{Method}

To uncover the reasoning of the model across its representations, we generate layer-wise explanations with CulTrace using the decoding question ``What is the assistant thinking about?'', referring to the model as the assistant, which gives us verbalised reasoning at that layer. This simple prompt works surprisingly well, as shown in Table \ref{tab:example}, following the simple prompts used in \citet{pan2026latentqa}. Layers 1-4 typically output reasoning that is not related to the question or culture, reflecting the pattern we saw in the evaluation of CulTrace (Section \ref{sec:eval_culturescope}). For Layer 5 onwards, we see that the model starts reasoning about the topic of the question and culture at large, eventually narrowing down to the answer. The model also goes down the wrong path, thinking about \textit{Japanese} snacks rather than \textit{Korean} ones before correcting itself and eventually getting to the right answer. These layer reasoning chains are incredibly useful for several analyses, allowing us to inspect what the model is focusing on and whether errors are due to a lack of cultural knowledge or overlooking certain parts of the question. They also mimic the intermediate steps for math, code or other related domains, allowing us to potentially design better process rewards for improvement in cultural reasoning. 

To better analyse these intermediate layer reasonings, we assign discrete labels to them, allowing for quantitative analysis. To design the labels, the authors qualitatively analysed reasoning chains and inductively created labels for each intermediate reasoning step. We also supplied several examples of reasoning chains to Claude Opus 4.8 in High Thinking Mode, prompting it to suggest labels. After manually reviewing each suggested label, we compared and integrated these suggestions, finally creating a taxonomy of labels across several dimensions (see Table~\ref{tab:taxonomy}).

We then prompt an LLM to annotate each intermediate reasoning based on labels from the taxonomy. The prompt (listed in App.\ref{appx:judge_prompt}) contains descriptions of labels, annotation guidelines, and examples. We use Qwen3.6-27B-FP8 \cite{qwen3.6-27b} as our LLM annotator. We evaluated the annotations on a randomly sampled set of 100 reasonings against two annotators, achieving an average accuracy of 96.9\% across all dimensions. We further report details about the human annotation, agreement, and accuracy in Appendix \ref{appx:judge_prompt}.

\begin{figure}[t]
    \centering
    \includegraphics[width=1\columnwidth]{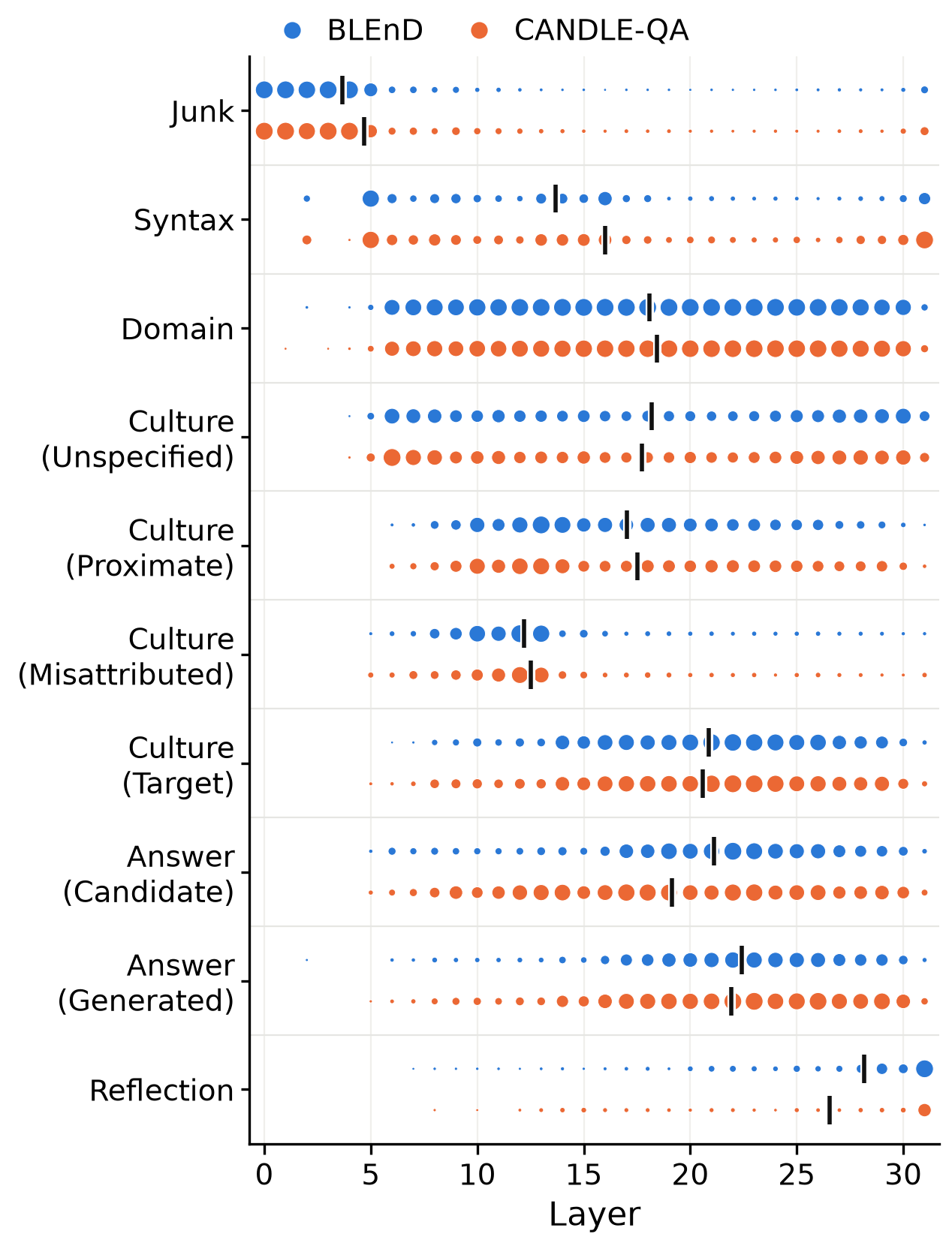}
    \caption{We show the layer position at which each dimension is annotated for the BLEnD dataset and CANDLE QA with \texttt{Llama-8b}. The size of the circle scales with the number of questions that are annotated for that label. The pipe symbol represents the mean layer index.}
    \label{fig:layer_tax_llama}
\end{figure}

\subsection{Results}\label{sec:tracing_result}

\paragraph{Characterisation of the Layers for Cultural Reasoning} Figure \ref{fig:layer_tax_llama} shows the layer position at which each dimension is annotated, averaged across questions, for BLEnD and CANDLE QA with \texttt{Llama-8b}. \textit{Overall, we see a clear order of intermediate reasoning steps; the model identifies the domain of the question and gradually resolves culture, then answers.} We find CulTrace generates irrelevant layer reasoning to culture early in the model, consistent with low reasoning recovery scores in those layers (\S\ref{sec:eval_culturescope}). The domain and syntax dimensions appear consistently across the model, while syntax appears less frequently. We also note that Reflection emerges in layers closer to the final generation, implying the model actively refines its answer after resolving other reasoning dimensions. Specifically, we observe that Culture (Misattributed) and Culture (Proximate) precede Culture (Target), suggesting that the model often focuses on proximate or wrong cultures in the early layers before arriving at the one the question is about in the later layers. Our results show a consistent trend on both BLEnD and CANDLE QA. Plots for \texttt{gemma-4b} and \texttt{Ministral-8b} show similar patterns, as in Figure \ref{fig:layer_tax_two} in the Appendix.

\begin{table}[]
\resizebox{\columnwidth}{!}{%
\begin{tabular}{@{}l|ccccc@{}}
\toprule
\multirow{4}{*}{\textbf{BLEnD}}     & \textbf{China}     & \textbf{US}     & \textbf{Spain}    & \textbf{S. Korea} & \textbf{Indonesia}   \\
& 12.8               & 13.2            & 13.9              & 15.5                 & 15.6                 \\ \cmidrule(l){2-6} 
& \textbf{Greece}    & \textbf{Iran}   & \textbf{Ethiopia} & \textbf{Azerbaijan}  & \textbf{Algeria}     \\ 
& 15.8               & 16.5            & 17.9              & 18.4                 & 21.9                 \\ \midrule
\multirow{4}{*}{\textbf{CANDLE QA}} & \textbf{US}        & \textbf{Spain}  & \textbf{China}    & \textbf{Greece}      & \textbf{S. Korea} \\ 
& 12.0               & 12.7            & 13.4              & 14.4                 & 14.6                 \\ \cmidrule(l){2-6} 
& \textbf{Indonesia} & \textbf{Brazil} & \textbf{Peru}     & \textbf{Iran}        & \textbf{Nepal}       \\ 
& 15.0               & 15.1            & 17.0              & 17.1                 & 17.4                 \\ \bottomrule
\end{tabular}%
}
\caption{We present the mean first layer that Culture (Target) appears within intermediate cultural reasoning across questions. The results are with \texttt{Llama-8b}.}
\label{tab:culture_tax}
\end{table}

\paragraph{Culture-level differences in Reasoning Process} To understand cross-cultural differences in how quickly a culture is resolved across model layers, we show the mean layer index of the first appearance that Culture (Target) is annotated for, across cultures in Table \ref{tab:culture_tax}. We find that China, the US, Spain, and S.Korea appear in the top-5 cultures on both datasets, suggesting that the model resolves target culture for those high-resourced cultures consistently earlier than remaining cultures. Underrepresented cultures like Algeria, Azerbaijan, and Nepal have the furthest first-layer indices, taking the longest to resolve. Table \ref{tab:layer_culture_reasoning} shows that ordering by the mean first layer of Culture (Target) is preserved with \texttt{gemma-4b} and \texttt{Ministral-8b}. \textit{Our result implies that the imbalanced encoding of cultural representation within the model parameters can be attributed to delayed target culture resolution with less-represented cultures.}

\section{Measuring Cultural Encoding}\label{sec:imbalance}

With intermediate cultural reasonings, we find that LLMs actively engage with culture in early-middle layers, reasoning about culture at a general level in the early layers, then pinning down to a specific culture, which can be the target culture or misattributed culture (Japan in Figure \ref{tab:example}). We next study which culture the model focuses on across layers, zooming in on this misattribution.

\subsection{Target Culture Identification}

To identify which culture the model is internally representing, we prompt the decoder model to verbalise the culture encoded in each layer's representation. We set the prompt to the decoder as ``What is the \textbf{culture} the assistant is thinking about and the topic of the question?''. Empirically, we find that additionally probing for the topic in the decoding question leads to better generation with CulTrace. To extract culture names from the sentences generated by CulTrace, we construct a dictionary of culture names to pattern match against. We initialise the dictionary with the original culture names and iteratively expand it by either adding new or related culture names that appear in the generations. We provide the dictionary and the parsing success rate in Appendix \ref{appx:parse}.

\begin{table}[]
\centering
\small
\begin{tabular}{rccc}
\toprule
\textbf{}     & \texttt{Llama-8b} & \texttt{Ministral-8b} & \texttt{gemma-4b} \\ \midrule
Spain         & \textbf{0.531}             & 0.388                 & 0.461             \\
Greece        & 0.527             & 0.319                 & 0.374             \\
China         & 0.506             & 0.360                 & \textbf{0.622}             \\
South Korea   & 0.498             & 0.322                 & 0.429             \\
Indonesia     & 0.476             & 0.371                 & 0.401             \\
Ethiopia      & 0.444             & 0.322                 & 0.344             \\
Iran          & 0.404             & 0.259                 & 0.450             \\
United States & 0.377             & \textbf{0.515}                 & 0.235             \\
Azerbaijan    & 0.218             & 0.095                 & 0.121             \\
Algeria       & \underline{0.069}             & \underline{0.026}                 & \underline{0.032}             \\ \bottomrule
\end{tabular}
\caption{We present averaged accuracies across layers for ten cultures with BLEnD. We mark the highest accuracy with \textbf{bold} and the lowest accuracy with \underline{underline}.}
\label{tab:acc_blend}
\end{table}

\subsection{Results}

\paragraph{Culture-Wise Accuracies}

Table \ref{tab:acc_blend} shows averaged accuracies across layers for target culture identification on the BLEnD dataset. We see that all models consistently have the lowest accuracies with Algeria and Azerbaijan. On the other hand, China and Spain rank within the top 3 with all models. Among models, \texttt{Llama-8b} shows the biggest variation across cultures and \texttt{gemma-4b} is the most consistent. Notably, \texttt{Llama-8b} and \texttt{gemma-4b} show low accuracies with the US, which implies that the models cannot verbalise the US when answering the cultural question about the US. This aligns with previous studies showing that the model defaults to the US when the culture is not specified in the prompt \cite{wang-etal-2024-countries, dai-etal-2025-word}.

\begin{table}[t]
    \centering
    \resizebox{\linewidth}{!}{
    \small
    \begin{tabular}{l|p{0.68\linewidth}}
    \toprule
        Target culture & Incorrectly predicted cultures (Top-5 Error)\\
    \midrule
        Algeria & Japan, North Africa, France, Morocco, Middle East \\
        Azerbaijan & Japan, Russia, Turkey, Kazakhstan, Caucasus \\
        China & Japan, Asia, Australia, US, South Korea \\
        Ethiopia  & Japan, Africa, South Africa, Kenya, US \\
        Greece  & Japan, Italy, Spain, US, UK \\
        Indonesia   & Japan, Asia, Australia, US, China \\
        Iran     & Japan, Middle East, Turkey, US, China \\
        South Korea & Japan, US, Asia, China, Australia \\
        Spain   & Japan, France, US, UK, Italy \\
        USA     & Japan, Western, UK, China, Canada \\
    \bottomrule
    \end{tabular}
    }
    \caption{Top-5 incorrectly predicted cultures for each target culture. The result is with \texttt{Llama-8b} on the BLEnD dataset.}
    \label{tab:bubble_table}
\end{table}

\paragraph{Error Analysis on Target Culture Identification}\label{sec:confusion}

We then look at cultures the LLM incorrectly predicts while processing cultural knowledge to answer the given question per target culture. Analysing incorrect culture predictions tells us what cultures are closely associated within the model representations. Table \ref{tab:bubble_table} shows the list of incorrectly predicted cultures for each culture with \texttt{Llama-8b} on the BLEnD dataset. As we see zero accuracy across all cultures until layer 5, we disregard those layers when aggregating occurrence frequencies. Results on other models and datasets are presented in Appendix \ref{appx:culture_acc}. We find a model-specific default culture (Japan for \texttt{Llama-8b}, US for \texttt{gemma-4b}) that the model erroneously outputs when answering questions about any culture. This mirrors the finding by \citet{delanda2026llmsobsessedjapaneseculture} that LLMs tend to refer to certain cultures (Japan, US, India) when they do not mention the target culture. 

\textit{We also observe that for most cultures, the model thinks about the broader geographical region before it reaches a specific culture.} For instance, for questions pertaining to Algeria, the model reasons about North Africa, for Iran, the Middle East and for the US, the West. This suggests that the model often reasons about the broader region of the world before pinning down the specific culture.

\section{Discussion \& Conclusion}

In this work, we propose a framework CulTrace to probe and elicit latent intermediate reasonings across the layers of a model when answering cultural questions. The method generalises to any task pertaining to LLMs, however in this study, we focus on cultural reasoning due to its challenging and subjective nature. The intermediate reasonings show how the model focuses on different parts of the cultural question across the layers - early layers focus on the syntax of the query and culture more broadly, while later layers focus on specific cultures, answer generation, and reflection.

Prior work simply localises cultural knowledge in deeper layers~\cite{namazifard-poech-2025-isolating, zou-etal-2026-deciphering} by showing that perturbing those layers brings the most performance degradation. Our findings extend this monolithic finding by further attributing specific reasoning roles to those layers. Those layers collaborate to resolve cultural knowledge through distinct and ordered stages of intermediate reasoning, comprised of domain engagement, target culture resolution, answer selection, and reflection (\S\ref{sec:tracing_result}).

At the same time, our results show that early layers play an important role that output-based evaluation cannot reveal. The model engages with the domain of the question before any culture is resolved, and it is in these earlier layers that broader or misattributed cultural context emerges ahead of the target culture. This broad-to-narrow trajectory is consistent with general MI findings, where early layers enrich representations with a wide set of related concepts and middle-to-later layers select those most relevant to the task~\cite{geva-etal-2023-dissecting, farahani-johansson-2024-deciphering, yu-etal-2024-mechanistic}. In the cultural setting, this means early layers activate a broad cultural neighbourhood, often a geographical region or a dominant culture, from which the specific target culture is only later pinned down. This narrowing is where cultural misattribution occurs: before resolving the target culture, LLMs often settle on a proximate one, reasoning about North Africa before Algeria or the Middle East before Iran, or falling back on a model-specific default (Japan for \texttt{Llama-8b}, the US for \texttt{gemma-4b}). Finally, we also see how higher-resourced cultures are resolved earlier in the layers but lower-represented ones are only resolved in the later layers, showing signs of potential cultural bias in how different cultures are processed internally.

These findings have clear implications for cultural alignment of LLMs. Because misattribution is established during the narrowing process rather than at the output, interventions may be most effective at the middle layers where the target culture is resolved, rather than at the final layers. The depth of culture resolution also differs across cultures, so targeted data augmentation or fine-tuning for less-represented cultures could aim specifically at strengthening mid-layer retrieval, where our results show the gap is largest. More broadly, tracing culture layer by layer offers a way to monitor whether an alignment intervention improves cultural balance internally, not only in the model's surface outputs. We intend to release the datasets of intermediate cultural reasonings as well as the trained decoder checkpoints for all models, which we hope will allow future research on interpretability of LLMs.

\section*{Limitations}

\paragraph{Dataset-related}
Although a growing number of benchmarks evaluate cultural knowledge in LLMs, few are suitable for our setup. To compare cultural knowledge across cultures, datasets must maintain a consistent QA format. As a result, our analyses inherit limitations of existing benchmarks, such as limited coverage of fine-grained cultures or subcultures. Nevertheless, our method is model- and task-agnostic and can be applied to any dataset that satisfies these requirements.

\paragraph{CulTrace-related}
Due to computational constraints, our experiments are limited to models of approximately 8B parameters. Even though we evaluate the faithfulness and domain transferability of CulTrace, further stress tests of these verbalisers are necessary.

\paragraph{Miscellaneous}
An AI writing assistant tool is used for a grammar check and stylistic improvement. None of the AI generations is used as is.

\section*{Acknowledgements}
$\begin{array}{l}\includegraphics[width=1cm]{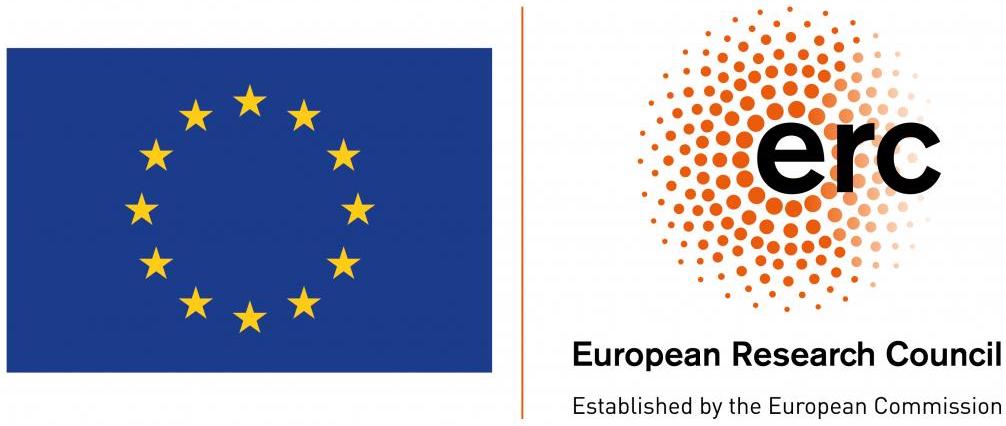} \end{array}$ 
This research was co-funded by the European Union (ERC, ExplainYourself, 101077481), the Carlsberg Foundation under grant number CF22-1461, a DFF Sapere
Aude research leader grant under grant agreement
No 0171-00034B, and supported by the Pioneer Centre for AI, DNRF grant number P1. It was also supported by Institute of Information \& communications Technology Planning \& Evaluation (IITP) grant funded by the Korea government (MSIT) (No. RS-2024-00509258 and No. RS-2024-00469482, Global AI Frontier Lab). 
Views and opinions expressed are however those of the author(s) only and do not necessarily reflect those of the European Union or the European Research Council. Neither the European Union nor the granting authority can be held responsible for them. We also thank Sarah Masud for their careful proof-reading.

\bibliography{custom}

@inproceedings{dai-etal-2025-word,
    title = "From Word to World: Evaluate and Mitigate Culture Bias in {LLM}s via Word Association Test",
    author = "Dai, Xunlian  and
      Zhou, Li  and
      Wang, Benyou  and
      Li, Haizhou",
    editor = "Christodoulopoulos, Christos  and
      Chakraborty, Tanmoy  and
      Rose, Carolyn  and
      Peng, Violet",
    booktitle = "Proceedings of the 2025 Conference on Empirical Methods in Natural Language Processing",
    month = nov,
    year = "2025",
    address = "Suzhou, China",
    publisher = "Association for Computational Linguistics",
    url = "https://aclanthology.org/2025.emnlp-main.1246/",
    doi = "10.18653/v1/2025.emnlp-main.1246",
    pages = "24510--24526",
    ISBN = "979-8-89176-332-6"
}

@inproceedings{wang-etal-2024-countries,
    title = "Not All Countries Celebrate Thanksgiving: On the Cultural Dominance in Large Language Models",
    author = "Wang, Wenxuan  and
      Jiao, Wenxiang  and
      Huang, Jingyuan  and
      Dai, Ruyi  and
      Huang, Jen-tse  and
      Tu, Zhaopeng  and
      Lyu, Michael",
    editor = "Ku, Lun-Wei  and
      Martins, Andre  and
      Srikumar, Vivek",
    booktitle = "Proceedings of the 62nd Annual Meeting of the Association for Computational Linguistics (Volume 1: Long Papers)",
    month = aug,
    year = "2024",
    address = "Bangkok, Thailand",
    publisher = "Association for Computational Linguistics",
    url = "https://aclanthology.org/2024.acl-long.345/",
    doi = "10.18653/v1/2024.acl-long.345",
    pages = "6349--6384"
}

@misc{qwen3.6-27b,
    title  = {{Qwen3.6-27B}: Flagship-Level Coding in a {27B} Dense Model},
    author = {{Qwen Team}},
    month  = {April},
    year   = {2026},
    url    = {https://qwen.ai/blog?id=qwen3.6-27b}
}

@misc{delanda2026llmsobsessedjapaneseculture,
      title={Why are all LLMs Obsessed with Japanese Culture? On the Hidden Cultural and Regional Biases of LLMs}, 
      author={Joseba Fernandez {De Landa} and Carla Perez-Almendros and Jose Camacho-Collados},
      year={2026},
      eprint={2604.21751},
      archivePrefix={arXiv},
      primaryClass={cs.CL},
      url={https://arxiv.org/abs/2604.21751}, 
}

@misc{khanuja2026steeringllmsculturallylocalized,
      title={Steering LLMs for Culturally Localized Generation}, 
      author={Simran Khanuja and Hongbin Liu and Shujian Zhang and John Lambert and Mingqing Chen and Rajiv Mathews and Lun Wang},
      year={2026},
      eprint={2603.23301},
      archivePrefix={arXiv},
      primaryClass={cs.CL},
      url={https://arxiv.org/abs/2603.23301}, 
}

@inproceedings{zou-etal-2026-deciphering,
    title = "Deciphering Cultural Representations in Large Language Models via Sparse Autoencoders",
    author = "Zou, Chenye  and
      Jiao, Difan  and
      Hu, Lijie",
    editor = "Liakata, Maria  and
      Moreira, Viviane P.  and
      Zhang, Jiajun  and
      Jurgens, David",
    booktitle = "Findings of the {A}ssociation for {C}omputational {L}inguistics: {ACL} 2026",
    month = jul,
    year = "2026",
    address = "San Diego, California, United States",
    publisher = "Association for Computational Linguistics",
    url = "https://aclanthology.org/2026.findings-acl.278/",
    doi = "10.18653/v1/2026.findings-acl.278",
    pages = "5656--5677",
    ISBN = "979-8-89176-395-1"
}

@inproceedings{veselovsky-etal-2026-localized,
    title = "Localized Cultural Knowledge is Conserved and Controllable in Large Language Models",
    author = "Veselovsky, Veniamin  and
      Arg{\i}n, Berke  and
      Stroebl, Benedikt  and
      Wendler, Chris  and
      West, Robert  and
      Evans, James  and
      Griffiths, Thomas L.  and
      Narayanan, Arvind",
    editor = "Liakata, Maria  and
      Moreira, Viviane P.  and
      Zhang, Jiajun  and
      Jurgens, David",
    booktitle = "Findings of the {A}ssociation for {C}omputational {L}inguistics: {ACL} 2026",
    month = jul,
    year = "2026",
    address = "San Diego, California, United States",
    publisher = "Association for Computational Linguistics",
    url = "https://aclanthology.org/2026.findings-acl.2141/",
    doi = "10.18653/v1/2026.findings-acl.2141",
    pages = "43152--43178",
    ISBN = "979-8-89176-395-1"
}

@inproceedings{namazifard-poech-2025-isolating,
    title = "Isolating Culture Neurons in Multilingual Large Language Models",
    author = "Namazifard, Danial  and
      Galke Poech, Lukas",
    editor = "Inui, Kentaro  and
      Sakti, Sakriani  and
      Wang, Haofen  and
      Wong, Derek F.  and
      Bhattacharyya, Pushpak  and
      Banerjee, Biplab  and
      Ekbal, Asif  and
      Chakraborty, Tanmoy  and
      Singh, Dhirendra Pratap",
    booktitle = "Proceedings of the 14th International Joint Conference on Natural Language Processing and the 4th Conference of the Asia-Pacific Chapter of the Association for Computational Linguistics",
    month = dec,
    year = "2025",
    address = "Mumbai, India",
    publisher = "The Asian Federation of Natural Language Processing and The Association for Computational Linguistics",
    url = "https://aclanthology.org/2025.findings-ijcnlp.45/",
    doi = "10.18653/v1/2025.findings-ijcnlp.45",
    pages = "768--785",
    ISBN = "979-8-89176-303-6"
}

@inproceedings{yu-etal-2024-mechanistic,
    title = "Mechanistic Understanding and Mitigation of Language Model Non-Factual Hallucinations",
    author = "Yu, Lei  and
      Cao, Meng  and
      Cheung, Jackie CK  and
      Dong, Yue",
    editor = "Al-Onaizan, Yaser  and
      Bansal, Mohit  and
      Chen, Yun-Nung",
    booktitle = "Findings of the Association for Computational Linguistics: EMNLP 2024",
    month = nov,
    year = "2024",
    address = "Miami, Florida, USA",
    publisher = "Association for Computational Linguistics",
    url = "https://aclanthology.org/2024.findings-emnlp.466/",
    doi = "10.18653/v1/2024.findings-emnlp.466",
    pages = "7943--7956"
}

@inproceedings{farahani-johansson-2024-deciphering,
    title = "Deciphering the Interplay of Parametric and Non-parametric Memory in Retrieval-augmented Language Models",
    author = "Farahani, Mehrdad  and
      Johansson, Richard",
    editor = "Al-Onaizan, Yaser  and
      Bansal, Mohit  and
      Chen, Yun-Nung",
    booktitle = "Proceedings of the 2024 Conference on Empirical Methods in Natural Language Processing",
    month = nov,
    year = "2024",
    address = "Miami, Florida, USA",
    publisher = "Association for Computational Linguistics",
    url = "https://aclanthology.org/2024.emnlp-main.943/",
    doi = "10.18653/v1/2024.emnlp-main.943",
    pages = "16966--16977"
}

@inproceedings{geva-etal-2023-dissecting,
    title = "Dissecting Recall of Factual Associations in Auto-Regressive Language Models",
    author = "Geva, Mor  and
      Bastings, Jasmijn  and
      Filippova, Katja  and
      Globerson, Amir",
    editor = "Bouamor, Houda  and
      Pino, Juan  and
      Bali, Kalika",
    booktitle = "Proceedings of the 2023 Conference on Empirical Methods in Natural Language Processing",
    month = dec,
    year = "2023",
    address = "Singapore",
    publisher = "Association for Computational Linguistics",
    url = "https://aclanthology.org/2023.emnlp-main.751/",
    doi = "10.18653/v1/2023.emnlp-main.751",
    pages = "12216--12235"
}

@misc{belrose2025tunedlens,
      title={Eliciting Latent Predictions from Transformers with the Tuned Lens}, 
      author={Nora Belrose and Igor Ostrovsky and Lev McKinney and Zach Furman and Logan Smith and Danny Halawi and Stella Biderman and Jacob Steinhardt},
      year={2025},
      eprint={2303.08112},
      archivePrefix={arXiv},
      primaryClass={cs.LG},
      url={https://arxiv.org/abs/2303.08112}, 
}

@inproceedings{geva-etal-2022-lm,
    title = "{LM}-Debugger: An Interactive Tool for Inspection and Intervention in Transformer-Based Language Models",
    author = "Geva, Mor  and
      Caciularu, Avi  and
      Dar, Guy  and
      Roit, Paul  and
      Sadde, Shoval  and
      Shlain, Micah  and
      Tamir, Bar  and
      Goldberg, Yoav",
    editor = "Che, Wanxiang  and
      Shutova, Ekaterina",
    booktitle = "Proceedings of the 2022 Conference on Empirical Methods in Natural Language Processing: System Demonstrations",
    month = dec,
    year = "2022",
    address = "Abu Dhabi, UAE",
    publisher = "Association for Computational Linguistics",
    url = "https://aclanthology.org/2022.emnlp-demos.2/",
    doi = "10.18653/v1/2022.emnlp-demos.2",
    pages = "12--21"
}

@inproceedings{
hernandez2024inspecting,
title={Inspecting and Editing Knowledge Representations in Language Models},
author={Evan Hernandez and Belinda Z. Li and Jacob Andreas},
booktitle={First Conference on Language Modeling},
year={2024},
url={https://openreview.net/forum?id=ADtL6fgNRv}
}

@inproceedings{zhou-etal-2025-mapo,
    title = "Does Mapo Tofu Contain Coffee? Probing {LLM}s for Food-related Cultural Knowledge",
    author = "Zhou, Li  and
      Karidi, Taelin  and
      Liu, Wanlong  and
      Garneau, Nicolas  and
      Cao, Yong  and
      Chen, Wenyu  and
      Li, Haizhou  and
      Hershcovich, Daniel",
    editor = "Chiruzzo, Luis  and
      Ritter, Alan  and
      Wang, Lu",
    booktitle = "Proceedings of the 2025 Conference of the Nations of the Americas Chapter of the Association for Computational Linguistics: Human Language Technologies (Volume 1: Long Papers)",
    month = apr,
    year = "2025",
    address = "Albuquerque, New Mexico",
    publisher = "Association for Computational Linguistics",
    url = "https://aclanthology.org/2025.naacl-long.496/",
    doi = "10.18653/v1/2025.naacl-long.496",
    pages = "9840--9867",
    ISBN = "979-8-89176-189-6"
}

@inproceedings{yu-etal-2024-revealing,
    title = "Revealing the Parametric Knowledge of Language Models: A Unified Framework for Attribution Methods",
    author = "Yu, Haeun  and
      Atanasova, Pepa  and
      Augenstein, Isabelle",
    editor = "Ku, Lun-Wei  and
      Martins, Andre  and
      Srikumar, Vivek",
    booktitle = "Proceedings of the 62nd Annual Meeting of the Association for Computational Linguistics (Volume 1: Long Papers)",
    month = aug,
    year = "2024",
    address = "Bangkok, Thailand",
    publisher = "Association for Computational Linguistics",
    url = "https://aclanthology.org/2024.acl-long.444/",
    doi = "10.18653/v1/2024.acl-long.444",
    pages = "8173--8186"
}

@article{cao-etal-2024-cultural,
    title = "Cultural Adaptation of Recipes",
    author = "Cao, Yong  and
      Kementchedjhieva, Yova  and
      Cui, Ruixiang  and
      Karamolegkou, Antonia  and
      Zhou, Li  and
      Dare, Megan  and
      Donatelli, Lucia  and
      Hershcovich, Daniel",
    journal = "Transactions of the Association for Computational Linguistics",
    volume = "12",
    year = "2024",
    address = "Cambridge, MA",
    publisher = "MIT Press",
    url = "https://aclanthology.org/2024.tacl-1.5/",
    doi = "10.1162/tacl_a_00634",
    pages = "80--99"
}

@inproceedings{wang-etal-2024-seaeval,
    title = "{S}ea{E}val for Multilingual Foundation Models: From Cross-Lingual Alignment to Cultural Reasoning",
    author = "Wang, Bin  and
      Liu, Zhengyuan  and
      Huang, Xin  and
      Jiao, Fangkai  and
      Ding, Yang  and
      Aw, AiTi  and
      Chen, Nancy",
    editor = "Duh, Kevin  and
      Gomez, Helena  and
      Bethard, Steven",
    booktitle = "Proceedings of the 2024 Conference of the North American Chapter of the Association for Computational Linguistics: Human Language Technologies (Volume 1: Long Papers)",
    month = jun,
    year = "2024",
    address = "Mexico City, Mexico",
    publisher = "Association for Computational Linguistics",
    url = "https://aclanthology.org/2024.naacl-long.22/",
    doi = "10.18653/v1/2024.naacl-long.22",
    pages = "370--390"
}

@inproceedings{naous-xu-2025-origin,
    title = "On The Origin of Cultural Biases in Language Models: From Pre-training Data to Linguistic Phenomena",
    author = "Naous, Tarek  and
      Xu, Wei",
    editor = "Chiruzzo, Luis  and
      Ritter, Alan  and
      Wang, Lu",
    booktitle = "Proceedings of the 2025 Conference of the Nations of the Americas Chapter of the Association for Computational Linguistics: Human Language Technologies (Volume 1: Long Papers)",
    month = apr,
    year = "2025",
    address = "Albuquerque, New Mexico",
    publisher = "Association for Computational Linguistics",
    url = "https://aclanthology.org/2025.naacl-long.326/",
    doi = "10.18653/v1/2025.naacl-long.326",
    pages = "6423--6443",
    ISBN = "979-8-89176-189-6"
}

@inproceedings{keleg-magdy-2023-dlama,
    title = "{DLAMA}: A Framework for Curating Culturally Diverse Facts for Probing the Knowledge of Pretrained Language Models",
    author = "Keleg, Amr  and
      Magdy, Walid",
    editor = "Rogers, Anna  and
      Boyd-Graber, Jordan  and
      Okazaki, Naoaki",
    booktitle = "Findings of the Association for Computational Linguistics: ACL 2023",
    month = jul,
    year = "2023",
    address = "Toronto, Canada",
    publisher = "Association for Computational Linguistics",
    url = "https://aclanthology.org/2023.findings-acl.389/",
    doi = "10.18653/v1/2023.findings-acl.389",
    pages = "6245--6266"
}

@inproceedings{meng2023locating,
author = {Meng, Kevin and Bau, David and Andonian, Alex and Belinkov, Yonatan},
title = {Locating and editing factual associations in GPT},
year = {2022},
isbn = {9781713871088},
publisher = {Curran Associates Inc.},
address = {Red Hook, NY, USA},
booktitle = {Proceedings of the 36th International Conference on Neural Information Processing Systems},
articleno = {1262},
numpages = {14},
location = {New Orleans, LA, USA},
series = {NIPS '22}
}

@article{hendryckstest2021,
      title={Measuring Massive Multitask Language Understanding},
      author={Dan Hendrycks and Collin Burns and Steven Basart and Andy Zou and Mantas Mazeika and Dawn Song and Jacob Steinhardt},
      journal={Proceedings of the International Conference on Learning Representations (ICLR)},
      year={2021}
    }

@inproceedings{pawar-etal-2025-presumed,
    title = "Presumed Cultural Identity: How Names Shape {LLM} Responses",
    author = "Pawar, Siddhesh Milind  and
      Arora, Arnav  and
      Kaffee, Lucie-Aim{\'e}e  and
      Augenstein, Isabelle",
    editor = "Christodoulopoulos, Christos  and
      Chakraborty, Tanmoy  and
      Rose, Carolyn  and
      Peng, Violet",
    booktitle = "Findings of the Association for Computational Linguistics: EMNLP 2025",
    month = nov,
    year = "2025",
    address = "Suzhou, China",
    publisher = "Association for Computational Linguistics",
    url = "https://aclanthology.org/2025.findings-emnlp.1207/",
    doi = "10.18653/v1/2025.findings-emnlp.1207",
    pages = "22147--22172",
    ISBN = "979-8-89176-335-7"
}

@inproceedings{
pan2026latentqa,
title={Latent{QA}: Teaching {LLM}s to Decode Activations Into Natural Language},
author={Alexander Pan and Lijie Chen and Jacob Steinhardt},
booktitle={The Fourteenth International Conference on Learning Representations},
year={2026},
url={https://openreview.net/forum?id=niUroX9EOd}
}

@misc{gemmateam2025,
      title={Gemma 3 Technical Report}, 
      author={Team Gemma-Team and Aishwarya Kamath and Johan Ferret and Shreya Pathak and Nino Vieillard and Ramona Merhej and Sarah Perrin and Tatiana Matejovicova and Alexandre Ramé and Morgane Rivière and Louis Rouillard and Thomas Mesnard and Geoffrey Cideron and Jean-bastien Grill and Sabela Ramos and Edouard Yvinec and Michelle Casbon and Etienne Pot and Ivo Penchev and Gaël Liu and Francesco Visin and Kathleen Kenealy and Lucas Beyer and Xiaohai Zhai and Anton Tsitsulin and Robert Busa-Fekete and Alex Feng and Noveen Sachdeva and Benjamin Coleman and Yi Gao and Basil Mustafa and Iain Barr and Emilio Parisotto and David Tian and Matan Eyal and Colin Cherry and Jan-Thorsten Peter and Danila Sinopalnikov and Surya Bhupatiraju and Rishabh Agarwal and Mehran Kazemi and Dan Malkin and Ravin Kumar and David Vilar and Idan Brusilovsky and Jiaming Luo and Andreas Steiner and Abe Friesen and Abhanshu Sharma and Abheesht Sharma and Adi Mayrav Gilady and Adrian Goedeckemeyer and Alaa Saade and Alex Feng and Alexander Kolesnikov and Alexei Bendebury and Alvin Abdagic and Amit Vadi and András György and André Susano Pinto and Anil Das and Ankur Bapna and Antoine Miech and Antoine Yang and Antonia Paterson and Ashish Shenoy and Ayan Chakrabarti and Bilal Piot and Bo Wu and Bobak Shahriari and Bryce Petrini and Charlie Chen and Charline Le Lan and Christopher A. Choquette-Choo and CJ Carey and Cormac Brick and Daniel Deutsch and Danielle Eisenbud and Dee Cattle and Derek Cheng and Dimitris Paparas and Divyashree Shivakumar Sreepathihalli and Doug Reid and Dustin Tran and Dustin Zelle and Eric Noland and Erwin Huizenga and Eugene Kharitonov and Frederick Liu and Gagik Amirkhanyan and Glenn Cameron and Hadi Hashemi and Hanna Klimczak-Plucińska and Harman Singh and Harsh Mehta and Harshal Tushar Lehri and Hussein Hazimeh and Ian Ballantyne and Idan Szpektor and Ivan Nardini and Jean Pouget-Abadie and Jetha Chan and Joe Stanton and John Wieting and Jonathan Lai and Jordi Orbay and Joseph Fernandez and Josh Newlan and Ju-yeong Ji and Jyotinder Singh and Kat Black and Kathy Yu and Kevin Hui and Kiran Vodrahalli and Klaus Greff and Linhai Qiu and Marcella Valentine and Marina Coelho and Marvin Ritter and Matt Hoffman and Matthew Watson and Mayank Chaturvedi and Michael Moynihan and Min Ma and Nabila Babar and Natasha Noy and Nathan Byrd and Nick Roy and Nikola Momchev and Nilay Chauhan and Noveen Sachdeva and Oskar Bunyan and Pankil Botarda and Paul Caron and Paul Kishan Rubenstein and Phil Culliton and Philipp Schmid and Pier Giuseppe Sessa and Pingmei Xu and Piotr Stanczyk and Pouya Tafti and Rakesh Shivanna and Renjie Wu and Renke Pan and Reza Rokni and Rob Willoughby and Rohith Vallu and Ryan Mullins and Sammy Jerome and Sara Smoot and Sertan Girgin and Shariq Iqbal and Shashir Reddy and Shruti Sheth and Siim Põder and Sijal Bhatnagar and Sindhu Raghuram Panyam and Sivan Eiger and Susan Zhang and Tianqi Liu and Trevor Yacovone and Tyler Liechty and Uday Kalra and Utku Evci and Vedant Misra and Vincent Roseberry and Vlad Feinberg and Vlad Kolesnikov and Woohyun Han and Woosuk Kwon and Xi Chen and Yinlam Chow and Yuvein Zhu and Zichuan Wei and Zoltan Egyed and Victor Cotruta and Minh Giang and Phoebe Kirk and Anand Rao and Kat Black and Nabila Babar and Jessica Lo and Erica Moreira and Luiz Gustavo Martins and Omar Sanseviero and Lucas Gonzalez and Zach Gleicher and Tris Warkentin and Vahab Mirrokni and Evan Senter and Eli Collins and Joelle Barral and Zoubin Ghahramani and Raia Hadsell and Yossi Matias and D. Sculley and Slav Petrov and Noah Fiedel and Noam Shazeer and Oriol Vinyals and Jeff Dean and Demis Hassabis and Koray Kavukcuoglu and Clement Farabet and Elena Buchatskaya and Jean-Baptiste Alayrac and Rohan Anil and Dmitry and Lepikhin and Sebastian Borgeaud and Olivier Bachem and Armand Joulin and Alek Andreev and Cassidy Hardin and Robert Dadashi and Léonard Hussenot},
      year={2025},
      eprint={2503.19786},
      archivePrefix={arXiv},
      primaryClass={cs.CL},
      url={https://arxiv.org/abs/2503.19786}, 
}

@inproceedings{candle2023,
  title={Extracting Cultural Commonsense Knowledge at Scale},
  author={Nguyen, Tuan-Phong and Razniewski, Simon and Varde, Aparna and Weikum, Gerhard},
  booktitle={Proceedings of the ACM Web Conference},
  year={2023}
}

@inproceedings{patchscope,
author = {Ghandeharioun, Asma and Caciularu, Avi and Pearce, Adam and Dixon, Lucas and Geva, Mor},
title = {Patchscopes: a unifying framework for inspecting hidden representations of language models},
year = {2024},
publisher = {JMLR.org},
booktitle = {Proceedings of the 41st International Conference on Machine Learning},
articleno = {620},
numpages = {25},
location = {Vienna, Austria},
series = {ICML'24}
}

@misc{grattafiori2024llama3herdmodels,
      title={The Llama 3 Herd of Models}, 
      author={Aaron Grattafiori and Abhimanyu Dubey and Abhinav Jauhri and Abhinav Pandey and Abhishek Kadian and Ahmad Al-Dahle and Aiesha Letman and Akhil Mathur and Alan Schelten and Alex Vaughan and Amy Yang and Angela Fan and Anirudh Goyal and Anthony Hartshorn and Aobo Yang and Archi Mitra and Archie Sravankumar and Artem Korenev and Arthur Hinsvark and Arun Rao and Aston Zhang and Aurelien Rodriguez and Austen Gregerson and Ava Spataru and Baptiste Roziere and Bethany Biron and Binh Tang and Bobbie Chern and Charlotte Caucheteux and Chaya Nayak and Chloe Bi and Chris Marra and Chris McConnell and Christian Keller and Christophe Touret and Chunyang Wu and Corinne Wong and Cristian Canton Ferrer and Cyrus Nikolaidis and Damien Allonsius and Daniel Song and Danielle Pintz and Danny Livshits and Danny Wyatt and David Esiobu and Dhruv Choudhary and Dhruv Mahajan and Diego Garcia-Olano and Diego Perino and Dieuwke Hupkes and Egor Lakomkin and Ehab AlBadawy and Elina Lobanova and Emily Dinan and Eric Michael Smith and Filip Radenovic and Francisco Guzmán and Frank Zhang and Gabriel Synnaeve and Gabrielle Lee and Georgia Lewis Anderson and Govind Thattai and Graeme Nail and Gregoire Mialon and Guan Pang and Guillem Cucurell and Hailey Nguyen and Hannah Korevaar and Hu Xu and Hugo Touvron and Iliyan Zarov and Imanol Arrieta Ibarra and Isabel Kloumann and Ishan Misra and Ivan Evtimov and Jack Zhang and Jade Copet and Jaewon Lee and Jan Geffert and Jana Vranes and Jason Park and Jay Mahadeokar and Jeet Shah and Jelmer van der Linde and Jennifer Billock and Jenny Hong and Jenya Lee and Jeremy Fu and Jianfeng Chi and Jianyu Huang and Jiawen Liu and Jie Wang and Jiecao Yu and Joanna Bitton and Joe Spisak and Jongsoo Park and Joseph Rocca and Joshua Johnstun and Joshua Saxe and Junteng Jia and Kalyan Vasuden Alwala and Karthik Prasad and Kartikeya Upasani and Kate Plawiak and Ke Li and Kenneth Heafield and Kevin Stone and Khalid El-Arini and Krithika Iyer and Kshitiz Malik and Kuenley Chiu and Kunal Bhalla and Kushal Lakhotia and Lauren Rantala-Yeary and Laurens van der Maaten and Lawrence Chen and Liang Tan and Liz Jenkins and Louis Martin and Lovish Madaan and Lubo Malo and Lukas Blecher and Lukas Landzaat and Luke de Oliveira and Madeline Muzzi and Mahesh Pasupuleti and Mannat Singh and Manohar Paluri and Marcin Kardas and Maria Tsimpoukelli and Mathew Oldham and Mathieu Rita and Maya Pavlova and Melanie Kambadur and Mike Lewis and Min Si and Mitesh Kumar Singh and Mona Hassan and Naman Goyal and Narjes Torabi and Nikolay Bashlykov and Nikolay Bogoychev and Niladri Chatterji and Ning Zhang and Olivier Duchenne and Onur Çelebi and Patrick Alrassy and Pengchuan Zhang and Pengwei Li and Petar Vasic and Peter Weng and Prajjwal Bhargava and Pratik Dubal and Praveen Krishnan and Punit Singh Koura and Puxin Xu and Qing He and Qingxiao Dong and Ragavan Srinivasan and Raj Ganapathy and Ramon Calderer and Ricardo Silveira Cabral and Robert Stojnic and Roberta Raileanu and Rohan Maheswari and Rohit Girdhar and Rohit Patel and Romain Sauvestre and Ronnie Polidoro and Roshan Sumbaly and Ross Taylor and Ruan Silva and Rui Hou and Rui Wang and Saghar Hosseini and Sahana Chennabasappa and Sanjay Singh and Sean Bell and Seohyun Sonia Kim and Sergey Edunov and Shaoliang Nie and Sharan Narang and Sharath Raparthy and Sheng Shen and Shengye Wan and Shruti Bhosale and Shun Zhang and Simon Vandenhende and Soumya Batra and Spencer Whitman and Sten Sootla and Stephane Collot and Suchin Gururangan and Sydney Borodinsky and Tamar Herman and Tara Fowler and Tarek Sheasha and Thomas Georgiou and Thomas Scialom and Tobias Speckbacher and Todor Mihaylov and Tong Xiao and Ujjwal Karn and Vedanuj Goswami and Vibhor Gupta and Vignesh Ramanathan and Viktor Kerkez and Vincent Gonguet and Virginie Do and Vish Vogeti and Vítor Albiero and Vladan Petrovic and Weiwei Chu and Wenhan Xiong and Wenyin Fu and Whitney Meers and Xavier Martinet and Xiaodong Wang and Xiaofang Wang and Xiaoqing Ellen Tan and Xide Xia and Xinfeng Xie and Xuchao Jia and Xuewei Wang and Yaelle Goldschlag and Yashesh Gaur and Yasmine Babaei and Yi Wen and Yiwen Song and Yuchen Zhang and Yue Li and Yuning Mao and Zacharie Delpierre Coudert and Zheng Yan and Zhengxing Chen and Zoe Papakipos and Aaditya Singh and Aayushi Srivastava and Abha Jain and Adam Kelsey and Adam Shajnfeld and Adithya Gangidi and Adolfo Victoria and Ahuva Goldstand and Ajay Menon and Ajay Sharma and Alex Boesenberg and Alexei Baevski and Allie Feinstein and Amanda Kallet and Amit Sangani and Amos Teo and Anam Yunus and Andrei Lupu and Andres Alvarado and Andrew Caples and Andrew Gu and Andrew Ho and Andrew Poulton and Andrew Ryan and Ankit Ramchandani and Annie Dong and Annie Franco and Anuj Goyal and Aparajita Saraf and Arkabandhu Chowdhury and Ashley Gabriel and Ashwin Bharambe and Assaf Eisenman and Azadeh Yazdan and Beau James and Ben Maurer and Benjamin Leonhardi and Bernie Huang and Beth Loyd and Beto De Paola and Bhargavi Paranjape and Bing Liu and Bo Wu and Boyu Ni and Braden Hancock and Bram Wasti and Brandon Spence and Brani Stojkovic and Brian Gamido and Britt Montalvo and Carl Parker and Carly Burton and Catalina Mejia and Ce Liu and Changhan Wang and Changkyu Kim and Chao Zhou and Chester Hu and Ching-Hsiang Chu and Chris Cai and Chris Tindal and Christoph Feichtenhofer and Cynthia Gao and Damon Civin and Dana Beaty and Daniel Kreymer and Daniel Li and David Adkins and David Xu and Davide Testuggine and Delia David and Devi Parikh and Diana Liskovich and Didem Foss and Dingkang Wang and Duc Le and Dustin Holland and Edward Dowling and Eissa Jamil and Elaine Montgomery and Eleonora Presani and Emily Hahn and Emily Wood and Eric-Tuan Le and Erik Brinkman and Esteban Arcaute and Evan Dunbar and Evan Smothers and Fei Sun and Felix Kreuk and Feng Tian and Filippos Kokkinos and Firat Ozgenel and Francesco Caggioni and Frank Kanayet and Frank Seide and Gabriela Medina Florez and Gabriella Schwarz and Gada Badeer and Georgia Swee and Gil Halpern and Grant Herman and Grigory Sizov and Guangyi and Zhang and Guna Lakshminarayanan and Hakan Inan and Hamid Shojanazeri and Han Zou and Hannah Wang and Hanwen Zha and Haroun Habeeb and Harrison Rudolph and Helen Suk and Henry Aspegren and Hunter Goldman and Hongyuan Zhan and Ibrahim Damlaj and Igor Molybog and Igor Tufanov and Ilias Leontiadis and Irina-Elena Veliche and Itai Gat and Jake Weissman and James Geboski and James Kohli and Janice Lam and Japhet Asher and Jean-Baptiste Gaya and Jeff Marcus and Jeff Tang and Jennifer Chan and Jenny Zhen and Jeremy Reizenstein and Jeremy Teboul and Jessica Zhong and Jian Jin and Jingyi Yang and Joe Cummings and Jon Carvill and Jon Shepard and Jonathan McPhie and Jonathan Torres and Josh Ginsburg and Junjie Wang and Kai Wu and Kam Hou U and Karan Saxena and Kartikay Khandelwal and Katayoun Zand and Kathy Matosich and Kaushik Veeraraghavan and Kelly Michelena and Keqian Li and Kiran Jagadeesh and Kun Huang and Kunal Chawla and Kyle Huang and Lailin Chen and Lakshya Garg and Lavender A and Leandro Silva and Lee Bell and Lei Zhang and Liangpeng Guo and Licheng Yu and Liron Moshkovich and Luca Wehrstedt and Madian Khabsa and Manav Avalani and Manish Bhatt and Martynas Mankus and Matan Hasson and Matthew Lennie and Matthias Reso and Maxim Groshev and Maxim Naumov and Maya Lathi and Meghan Keneally and Miao Liu and Michael L. Seltzer and Michal Valko and Michelle Restrepo and Mihir Patel and Mik Vyatskov and Mikayel Samvelyan and Mike Clark and Mike Macey and Mike Wang and Miquel Jubert Hermoso and Mo Metanat and Mohammad Rastegari and Munish Bansal and Nandhini Santhanam and Natascha Parks and Natasha White and Navyata Bawa and Nayan Singhal and Nick Egebo and Nicolas Usunier and Nikhil Mehta and Nikolay Pavlovich Laptev and Ning Dong and Norman Cheng and Oleg Chernoguz and Olivia Hart and Omkar Salpekar and Ozlem Kalinli and Parkin Kent and Parth Parekh and Paul Saab and Pavan Balaji and Pedro Rittner and Philip Bontrager and Pierre Roux and Piotr Dollar and Polina Zvyagina and Prashant Ratanchandani and Pritish Yuvraj and Qian Liang and Rachad Alao and Rachel Rodriguez and Rafi Ayub and Raghotham Murthy and Raghu Nayani and Rahul Mitra and Rangaprabhu Parthasarathy and Raymond Li and Rebekkah Hogan and Robin Battey and Rocky Wang and Russ Howes and Ruty Rinott and Sachin Mehta and Sachin Siby and Sai Jayesh Bondu and Samyak Datta and Sara Chugh and Sara Hunt and Sargun Dhillon and Sasha Sidorov and Satadru Pan and Saurabh Mahajan and Saurabh Verma and Seiji Yamamoto and Sharadh Ramaswamy and Shaun Lindsay and Shaun Lindsay and Sheng Feng and Shenghao Lin and Shengxin Cindy Zha and Shishir Patil and Shiva Shankar and Shuqiang Zhang and Shuqiang Zhang and Sinong Wang and Sneha Agarwal and Soji Sajuyigbe and Soumith Chintala and Stephanie Max and Stephen Chen and Steve Kehoe and Steve Satterfield and Sudarshan Govindaprasad and Sumit Gupta and Summer Deng and Sungmin Cho and Sunny Virk and Suraj Subramanian and Sy Choudhury and Sydney Goldman and Tal Remez and Tamar Glaser and Tamara Best and Thilo Koehler and Thomas Robinson and Tianhe Li and Tianjun Zhang and Tim Matthews and Timothy Chou and Tzook Shaked and Varun Vontimitta and Victoria Ajayi and Victoria Montanez and Vijai Mohan and Vinay Satish Kumar and Vishal Mangla and Vlad Ionescu and Vlad Poenaru and Vlad Tiberiu Mihailescu and Vladimir Ivanov and Wei Li and Wenchen Wang and Wenwen Jiang and Wes Bouaziz and Will Constable and Xiaocheng Tang and Xiaojian Wu and Xiaolan Wang and Xilun Wu and Xinbo Gao and Yaniv Kleinman and Yanjun Chen and Ye Hu and Ye Jia and Ye Qi and Yenda Li and Yilin Zhang and Ying Zhang and Yossi Adi and Youngjin Nam and Yu and Wang and Yu Zhao and Yuchen Hao and Yundi Qian and Yunlu Li and Yuzi He and Zach Rait and Zachary DeVito and Zef Rosnbrick and Zhaoduo Wen and Zhenyu Yang and Zhiwei Zhao and Zhiyu Ma},
      year={2024},
      eprint={2407.21783},
      archivePrefix={arXiv},
      primaryClass={cs.AI},
      url={https://arxiv.org/abs/2407.21783}, 
}

@article{myung2024blend,
  title={Blend: A benchmark for llms on everyday knowledge in diverse cultures and languages},
  author={Myung, Junho and Lee, Nayeon and Zhou, Yi and Jin, Jiho and Putri, Rifki and Antypas, Dimosthenis and Borkakoty, Hsuvas and Kim, Eunsu and Perez-Almendros, Carla and Ayele, Abinew Ali and others},
  journal={Advances in Neural Information Processing Systems},
  volume={37},
  pages={78104--78146},
  year={2024}
}

@article{pawar2025survey,
    author = {Pawar, Siddhesh and Park, Junyeong and Jin, Jiho and Arora, Arnav and Myung, Junho and Yadav, Srishti and Haznitrama, Faiz Ghifari and Song, Inhwa and Oh, Alice and Augenstein, Isabelle},
    title = {Survey of Cultural Awareness in Language Models: Text and Beyond},
    journal = {Computational Linguistics},
    pages = {1-98},
    year = {2025},
    month = {07},
    issn = {0891-2017},
    doi = {10.1162/COLI.a.14},
    url = {https://doi.org/10.1162/COLI.a.14},
    eprint = {https://direct.mit.edu/coli/article-pdf/doi/10.1162/COLI.a.14/2523159/coli.a.14.pdf},
}

@inproceedings{hasan-etal-2025-nativqa,
    title = "{N}ativ{QA}: Multilingual Culturally-Aligned Natural Query for {LLM}s",
    author = "Hasan, Md Arid  and
      Hasanain, Maram  and
      Ahmad, Fatema  and
      Laskar, Sahinur Rahman  and
      Upadhyay, Sunaya  and
      Sukhadia, Vrunda N  and
      Kutlu, Mucahid  and
      Chowdhury, Shammur Absar  and
      Alam, Firoj",
    editor = "Che, Wanxiang  and
      Nabende, Joyce  and
      Shutova, Ekaterina  and
      Pilehvar, Mohammad Taher",
    booktitle = "Findings of the Association for Computational Linguistics: ACL 2025",
    month = jul,
    year = "2025",
    address = "Vienna, Austria",
    publisher = "Association for Computational Linguistics",
    url = "https://aclanthology.org/2025.findings-acl.770/",
    pages = "14886--14909",
    ISBN = "979-8-89176-256-5",
}
\bibliographystyle{acl_natbib}

\clearpage
\appendix

\begin{table*}[t!]
\resizebox{\textwidth}{!}{%
\begin{tabular}{@{}c|c|c|l|l@{}}
\toprule
       Dataset & \textbf{Task}           & \multicolumn{1}{l|}{\textbf{Number of Questions}} & \multicolumn{1}{c|}{\textbf{Domain (\# of Qs)}}                                                                              & \multicolumn{1}{c}{\textbf{List of Cultures}}                                                                                                                                                                                                                  \\ \midrule
\begin{tabular}[c]{@{}c@{}}BLEnD\\\citet{myung2024blend}\end{tabular}   & \begin{tabular}[c]{@{}c@{}}Cultural\\Commonsense QA\end{tabular} & 4,020                                              & \begin{tabular}[c]{@{}l@{}}Education (488) \\ Food (979) \\ Holidays/Celebration/Leisure (816)\\ Sport (816)\\Work life (452)\\Family (469)\end{tabular} & \begin{tabular}[c]{@{}l@{}}Africa: Algeria, Ethiopia\\ Europe: Spain, Greece\\ North America: United States \\ East Asia: China, South Korea\\ South Asia: Indonesia \\ West Asia: Iran, Azerbaijan\end{tabular} \\ 

\midrule

\begin{tabular}[c]{@{}c@{}}CANDLE QA\\\citet{candle2023}\end{tabular} & Cultural knowledge QA           & 1,916                                              & \begin{tabular}[c]{@{}l@{}}Food (585) \\ Traditions (516) \\ Rituals (406)\\ Drink (322)\\ Clothing (87) \end{tabular} & \begin{tabular}[c]{@{}l@{}} Europe: Spain, Greece\\ North America: United States \\ South America: Brazil, Peru \\ East Asia: China, South Korea\\ South Asia: Indonesia, Nepal\\ West Asia: Iran \end{tabular}                        \\ \bottomrule
\end{tabular}%
}
\caption{Details for the datasets}
\label{tab:datasets}
\end{table*}

\section{Dataset}

\subsection{Dataset Details}\label{appendix:dataset}

In our experiments, we utilise BLEnD~\cite{myung2024blend} and CANDLE QA~\cite{candle2023}. BLEnD comprises 500 short-answer question-answer pairs for each culture, where the answers vary depending on the cultural or regional context. The filtered version of CANDLE from \citet{pawar-etal-2025-presumed} contains 23,421 assertions, with five domains. To reduce computational cost, we select ten cultures from BLEnD and CANDLE QA, ensuring equal distribution across the globe. For CANDLE QA, we also sample 40 assertions from each facet for each culture, and for cultures that do not have 200 assertions, we include all assertions. As a result, 1,967 assertions are selected. We provide brief data statistics and dataset characteristics in Table~\ref{tab:datasets}.

\subsection{Creation of CANDLE QA}\label{appx:cloze_to_qa}

As CANDLE is not inherently in a QA format, we convert those assertions into questions having annotated concepts as gold answers. Before conversion, we remove assertions that do not have the annotated concept verbatim in their assertion, which leaves us 1,916 assertions. Authors manually inspect converted questions to determine whether the LLM's conversion is coherent and does not contain a gold answer. The prompt we use for the conversion is provided in Figure \ref{fig:convert_prompt}. Along with an assertion, the model is given the target culture, domain, and the concept.

\section{Training of CulTrace}\label{appendix:culturescope}

\subsection{Training details}\label{appx:training_detail}

We train 102 LoRA adapters to obtain decoder models for each layer with three different LLMs. We train all models for one epoch, with the training data from LatentQA. For \texttt{Llama-8b}, we adapt training hyperparameters from \citet{pan2026latentqa}: batch size 8, 16 gradient accumulation steps, learning rate 1e-4, LoRA rank 32 and alpha 64. For \texttt{gemma-4b} and \texttt{Ministral-8b}, we perform a small sweep for training hyperparameter searching and track the evaluation loss with 10\% of the training dataset. We train \texttt{gemma-4b} with batch size 8, 16 gradient accumulation steps, learning rate 5e-5, LoRA rank 32 and alpha 64. \texttt{Ministral-8b} is trained with batch size 2, 64 gradient accumulation steps, learning rate 5e-5, LoRA rank 32 and alpha 64. Training of one LoRA adapter took 1.5 hours on one Nvidia B200 gpu.

\subsection{Evaluation of Trained Models}\label{appx:loss}

\begin{figure}
    \centering
    \includegraphics[width=1\columnwidth]{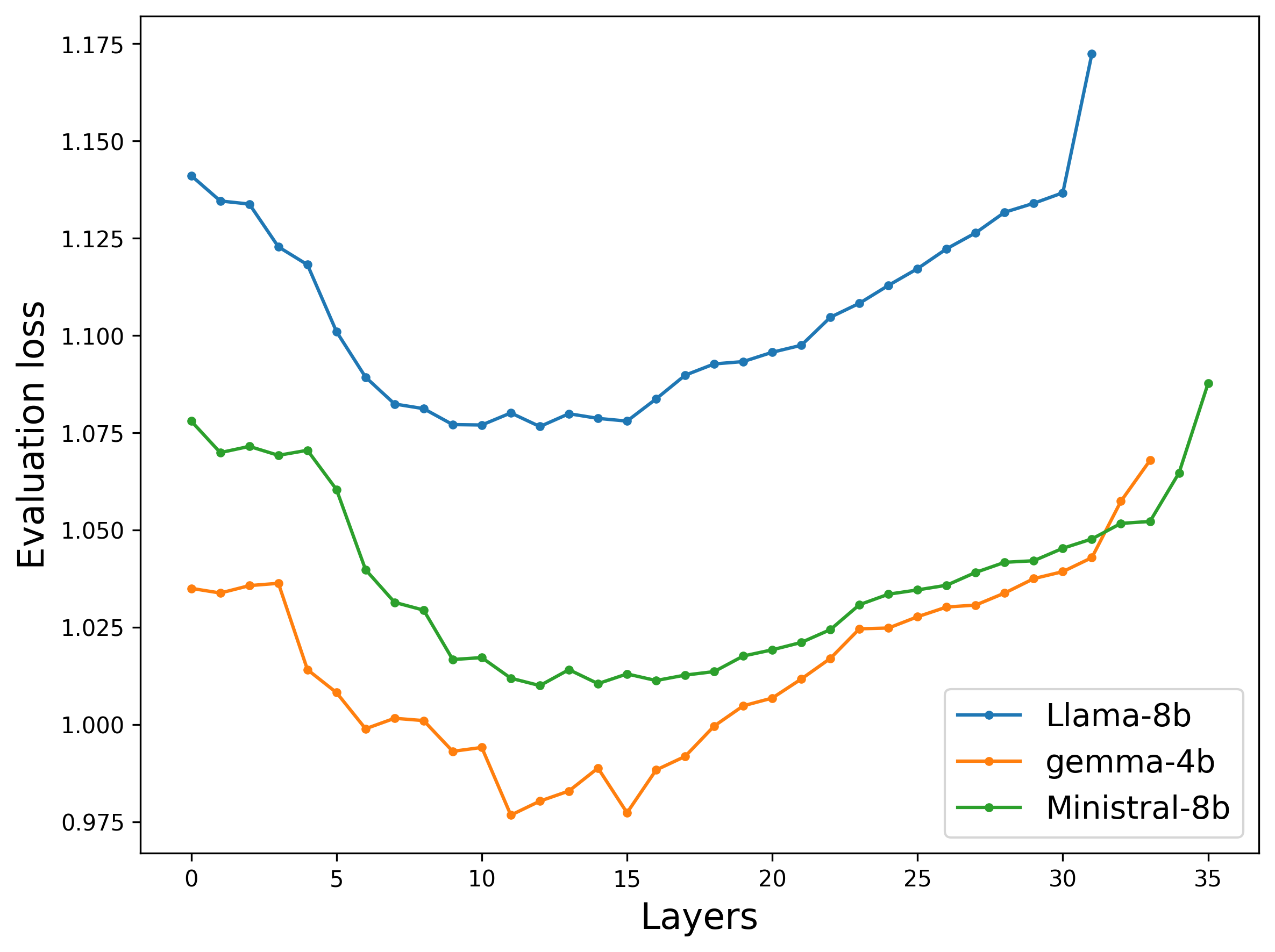}
    \caption{We present losses from the trained model on the test set from LatentQA.}
    \label{fig:eval_loss}
\end{figure}

We train one model per layer, and evaluate in two ways: a test set from LatentQA and attribute extraction as in Patchscopes \cite{patchscope} and LatentQA. Figure \ref{fig:eval_loss} shows evaluation loss on the test set from each layer. We see lower evaluation losses on middle layers.

\begin{figure}
    \centering
    \includegraphics[width=1\columnwidth]{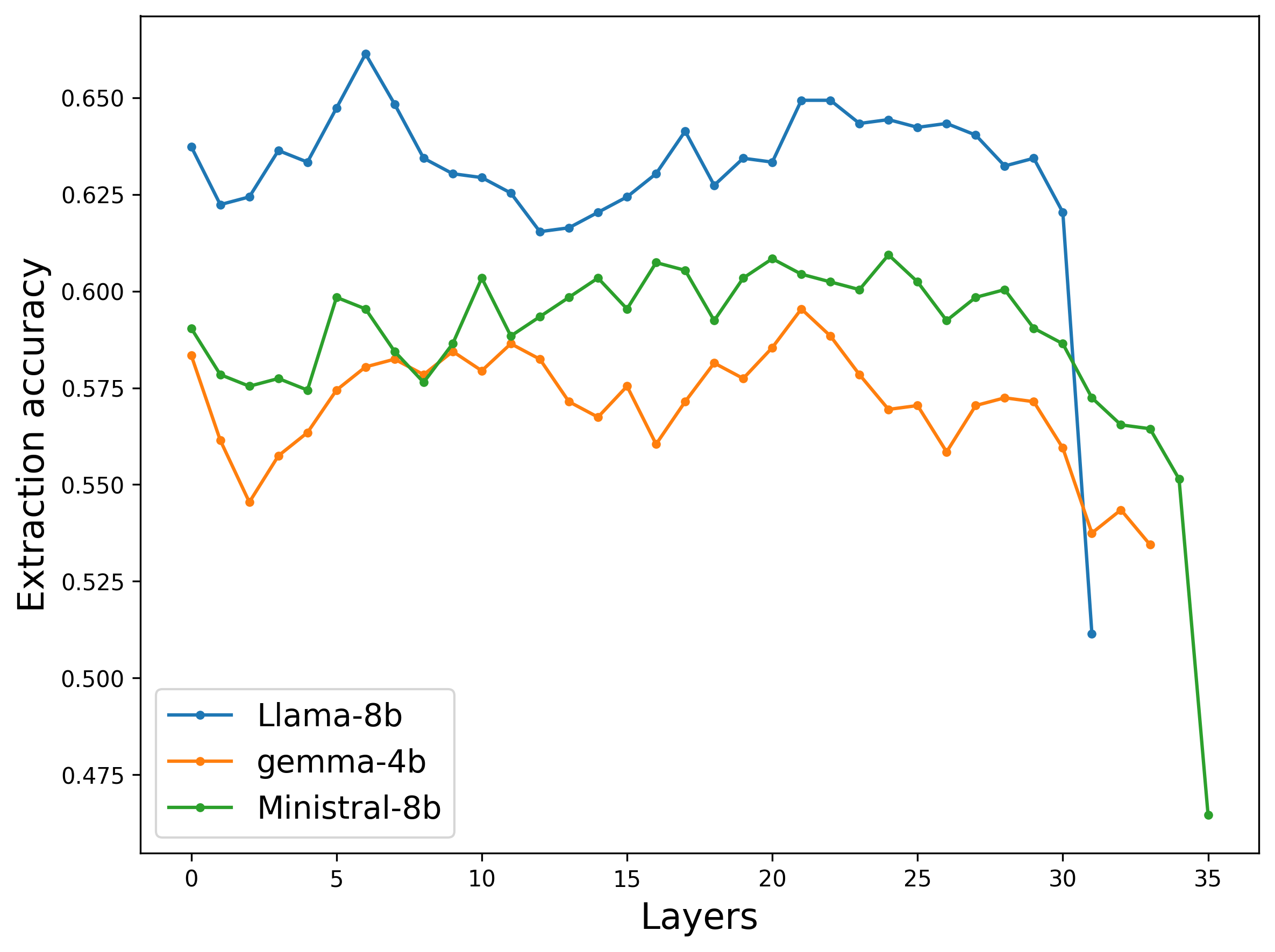}
    \caption{We present relational knowledge extraction accuracies from the trained model on the test set from LatentQA.}
    \label{fig:eval_acc}
\end{figure}

We also evaluate trained models on the relational knowledge extraction task as in LatentQA. In the relational knowledge extraction task, given hidden representations of a sentence containing a subject (e.g., ``The official currency of the United States''), the decoder model is asked to answer the decoding question (e.g., ``What is the currency of the country?''). We use the evaluation dataset from Patchscopes \cite{patchscope}. Figure \ref{fig:eval_acc} shows relational knowledge extraction accuracies from each layer with all models. We see that, for each model, accuracies vary by less than 0.03.

We release the LoRA fine-tuned adapter weights publicly (\texttt{gemma-4b}\footnote{\url{https://huggingface.co/copenlu/CulTrace-latentqa-decoder-gemma-3-4b-it}}, \texttt{llama-8b}\footnote{\url{https://huggingface.co/copenlu/CulTrace-latentqa-decoder-llama-3-8b-instruct}}, \texttt{Ministral-8b}\footnote{\url{https://huggingface.co/copenlu/CulTrace-latentqa-decoder-ministral-8b-instruct}})

\begin{figure*}
     \centering
     \begin{subfigure}[b]{0.47\textwidth}
         \centering
         \includegraphics[width=\columnwidth]{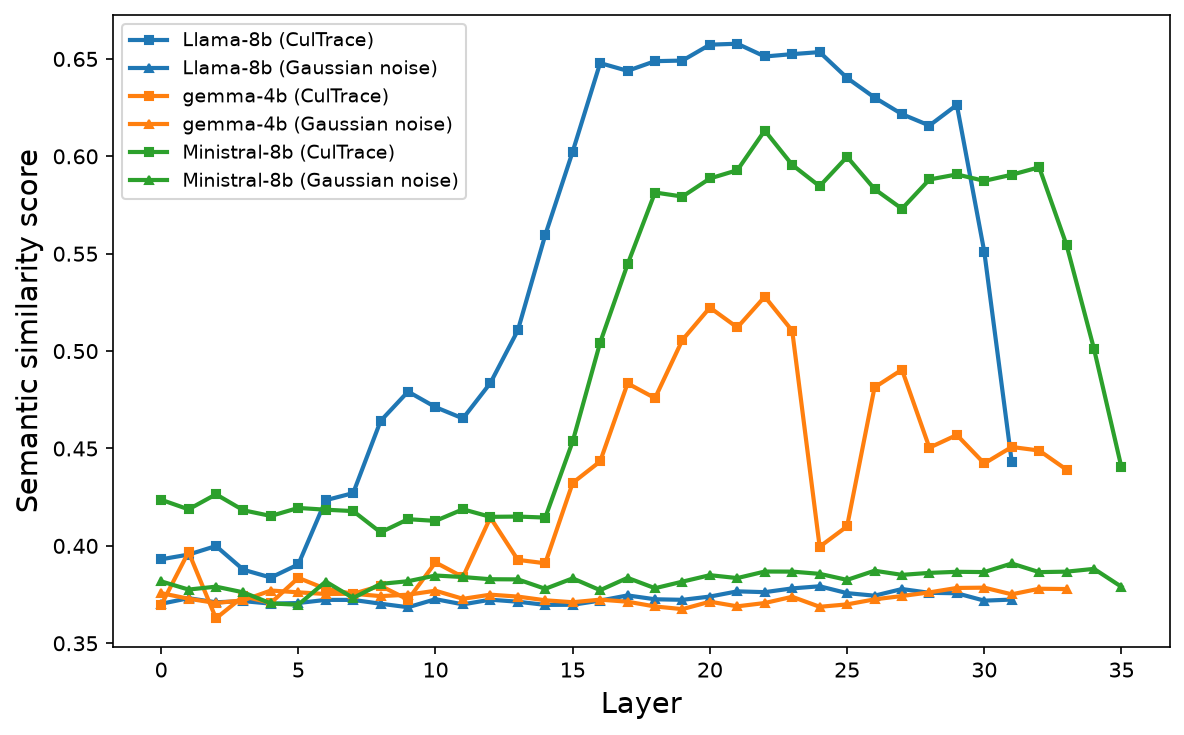}
         \caption{Comparison between CulTrace and patching Gaussian noise vectors}
     \end{subfigure}
     \hfill
     \begin{subfigure}[b]{0.47\textwidth}
         \centering
         \includegraphics[width=\columnwidth]{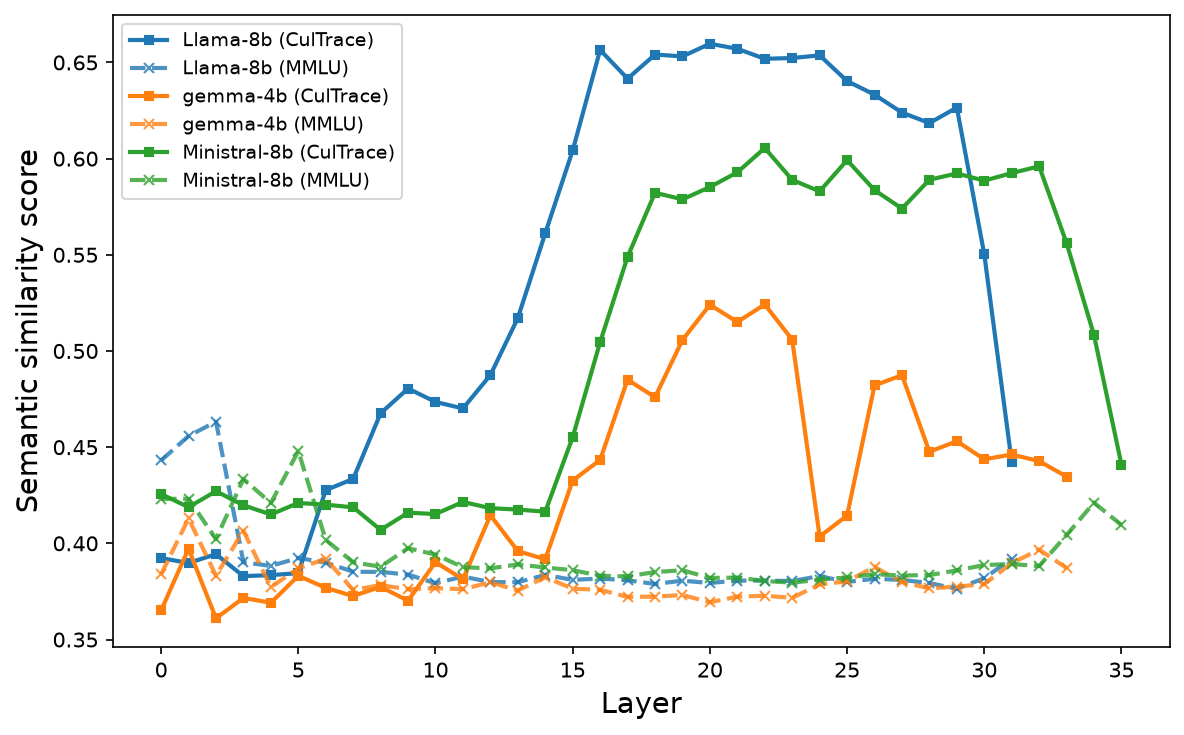}
         \caption{Comparison between cultural questions from the CANDLE QA and science questions from MMLU}
     \end{subfigure}
        \caption{We present semantic similarity scores with different set-ups with CANDLE QA.}
        \label{fig:candle_recovery}
\end{figure*}

\section{Evaluation of CulTrace}\label{appendix:culturescope_eval}

\subsection{Comparison with Gaussian Noise Vector}\label{appendix:noise}

\begin{figure}[t]
    \centering
    \includegraphics[width=1\columnwidth]{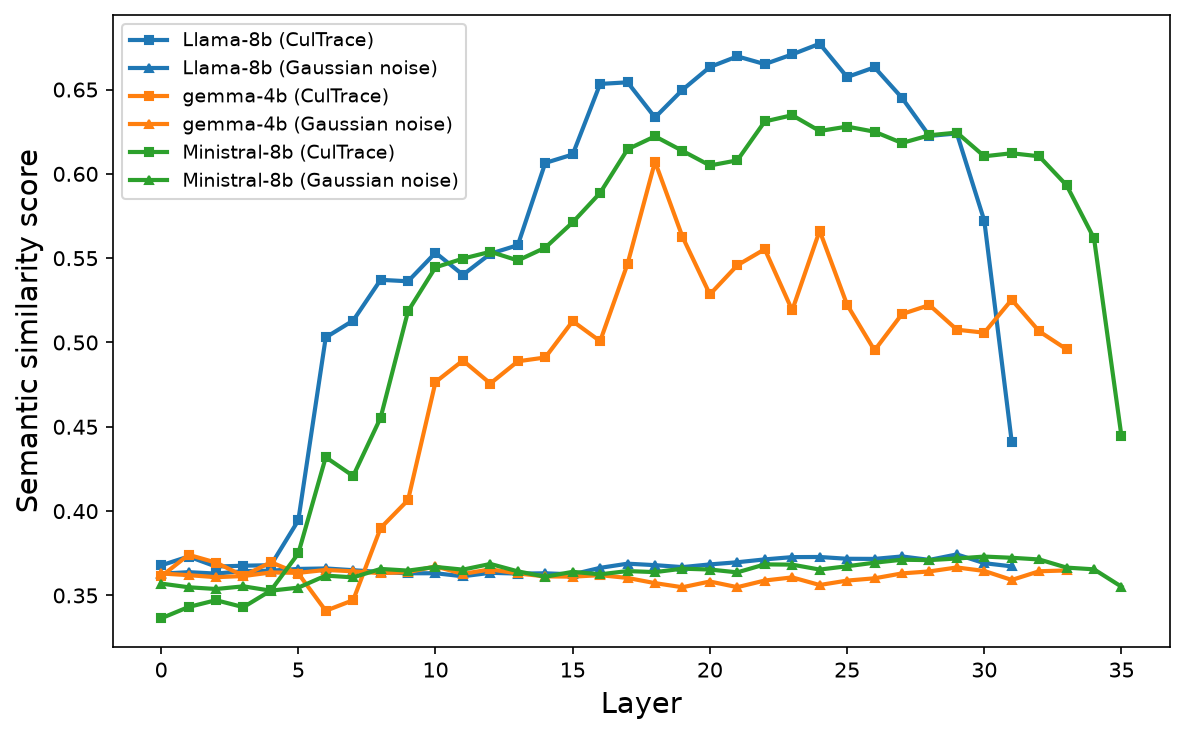}
    \caption{We present semantic similarity scores from CulTrace and patching a Gaussian noise vector with the BLEnD dataset.}
    \label{fig:noise_blend}
\end{figure}

We compare the semantic similarity of CulTrace across layers with a Gaussian noise vector, which serves as a dummy vector. Higher semantic similarity scores than Gaussian noise vectors show that culturally relevant outputs from CulTrace are decoded from patched representations. Figure \ref{fig:noise_blend} and Figure \ref{fig:candle_recovery} show the layer-wise semantic similarity scores with different set-ups.

\begin{figure*}
     \centering
     \begin{subfigure}[b]{0.47\textwidth}
         \centering
         \includegraphics[width=\columnwidth]{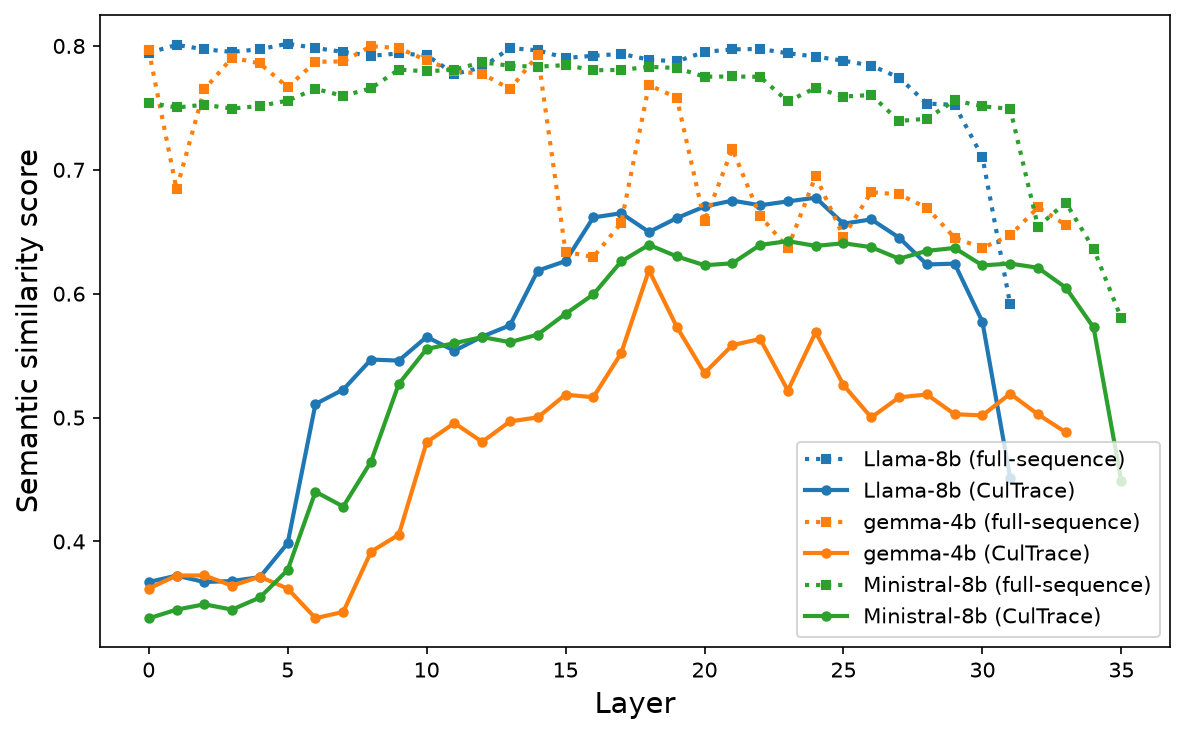}
         \caption{Results with the BLEnD dataset}
     \end{subfigure}
     \hfill
     \begin{subfigure}[b]{0.47\textwidth}
         \centering
         \includegraphics[width=\columnwidth]{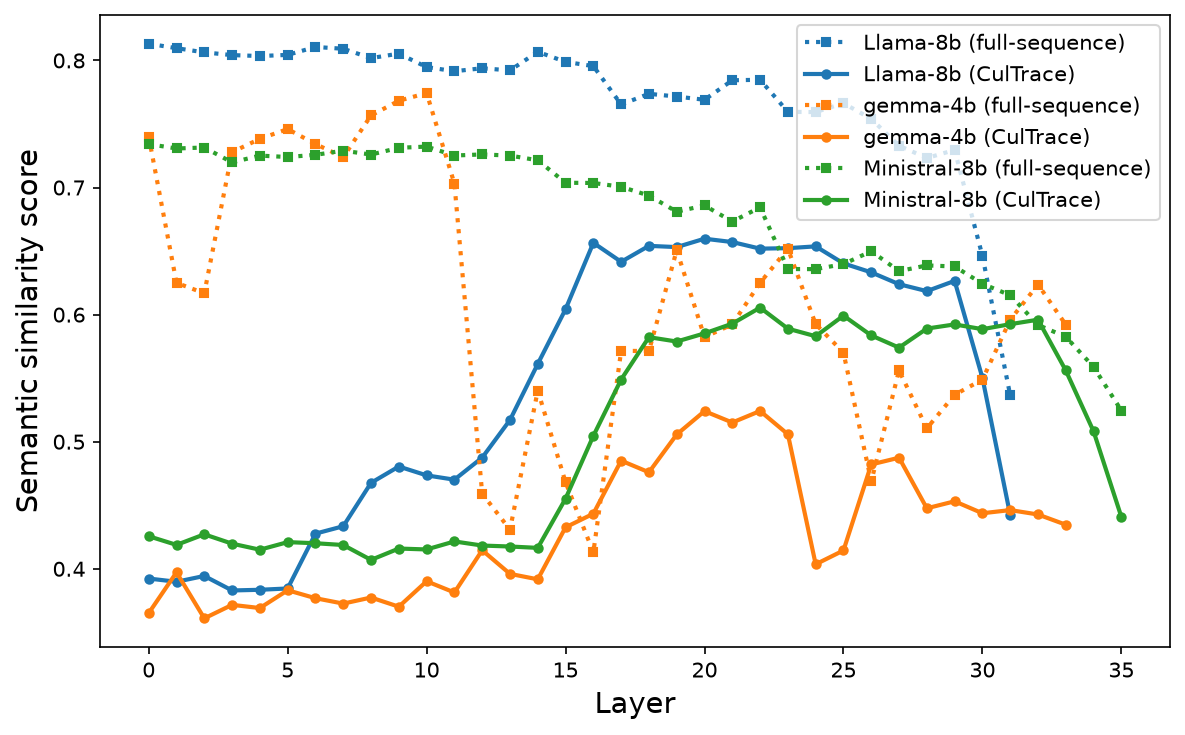}
         \caption{Results with CANDLE QA}
     \end{subfigure}
        \caption{We compare semantic similarity scores across layers between CulTrace and extracting representations from the full sequence (full-sequence).}
        \label{fig:compare_full}
\end{figure*}

\subsection{Position of Representation Extraction}\label{eval:pos}

With CulTrace, we extract the representation at the position of the instruction $\mathcal{I}$ rather than the full input sequence to ensure that the decoder model decodes the model's reasoning toward $Y$. Extracting representations from the full sequence, which includes the question $q$, risks introducing a confound: the decoder LLM may simply read and answer the question directly rather than decode the target LLM's internal reasoning. We therefore treat full-sequence extraction as an upper bound on representation decoding performance and compare it against CulTrace to verify that our instruction extraction captures the target LLM's reasoning without relying on direct access to the question. We expect a small gap between the semantic similarity scores from CulTrace and those from full-sequence. Processing 200 questions takes two hours on one partition of Nvidia B200 gpu (1/7 of the entire gpu).

Figure \ref{fig:compare_full} presents layer-wise semantic similarity scores for the two extraction settings on the BLEnD dataset. Full-sequence extraction achieves consistently high semantic similarity scores with the model outputs since the decoder model answers the question directly. We find that the gap between CulTrace and the full-sequence upper bound is smallest in the middle-to-late layers and largest in the early layers.

\section{LLM Annotator Details}

\subsection{Prompt}\label{appx:judge_prompt}

To develop the prompt, the authors prompted Claude Opus 4.8 to generate a seed prompt for labelling with the provided taxonomy. The authors then manually revised the prompt, elaborating each dimension, fixing the output formatting, and adding annotation guidelines for additional context. This was followed by repeated iterations of evaluation on a validation set of 50 reasonings, which were qualitatively assessed for correctness. Based on the above, the authors revised the prompt by adding further annotation guidelines or more examples. In addition to the prompt, the LLM is provided with the intermediate reasoning, along with the question, target culture, and the model's answer, for additional context, which helps produce the labels. 

\subsection{Human Evaluation}

\begin{table}[]
    \centering
    \begin{tabular}{rcc}
    \toprule
    Dimension & Agreement & Kappa\\
    \midrule
    Culture    & 0.920 & 0.861 \\
    Domain     & 0.970 & 0.930 \\
    Answer     & 0.960 & 0.483 \\
    Reflection & 0.970 & 0.556 \\
    Syntax     & 0.990 & 0.000 \\
    Junk       & 0.940 & 0.844 \\
    \midrule
    Mean       & 0.958 & 0.613\\
    \bottomrule
    \end{tabular}
    \caption{Agreement statistics between the two annotators for the human evaluation. Agreement reflects the percentage of samples where labels are an exact match. Kappa represents Cohen's kappa score for all samples. Syntax Kappa is 0 since one of the annotators did not output any labels for syntax.}
    \label{tab:human_agreement}
\end{table}

\begin{table}[]
\centering
\begin{tabular}{rcc}
\toprule
Dimension & Accuracy & N\_consensus \\
\midrule
Culture    & 0.989 & 92 \\
Domain     & 0.979 & 97 \\
Answer     & 0.979 & 96 \\
Reflection & 0.990 & 97 \\
Syntax     & 0.909 & 99 \\
Junk       & 0.968 & 94 \\
\midrule
Mean       & 0.969 & 95.833 \\
\bottomrule
\end{tabular}
\caption{LLM Annotator accuracies per dimension for the gold subset where both human annotators have consensus over the labels. N\_consensus represents the number of samples (out of 100) with consensus, over which the LLM was evaluated.}
\label{tab:annotation_acc}
\end{table}

To evaluate LLM annotations, two authors annotate 100 reasonings from the CulTrace outputs. Table \ref{tab:human_agreement} shows agreement statistics between the two annotators. Human evaluation achieves 0.95 agreement and 0.613 Cohen's Kappa across all dimensions on average, demonstrating substantial agreement. Table \ref{tab:annotation_acc} further shows LLM Annotator accuracies per dimension, measured across the instances with consensus between the two annotators. Accuracy of the LLM is high across all dimensions, with scores varying from 90\% for syntax to 99\% for Reflection.

\begin{figure*}
     \centering
     \begin{subfigure}[b]{0.47\textwidth}
         \centering
         \includegraphics[width=\columnwidth]{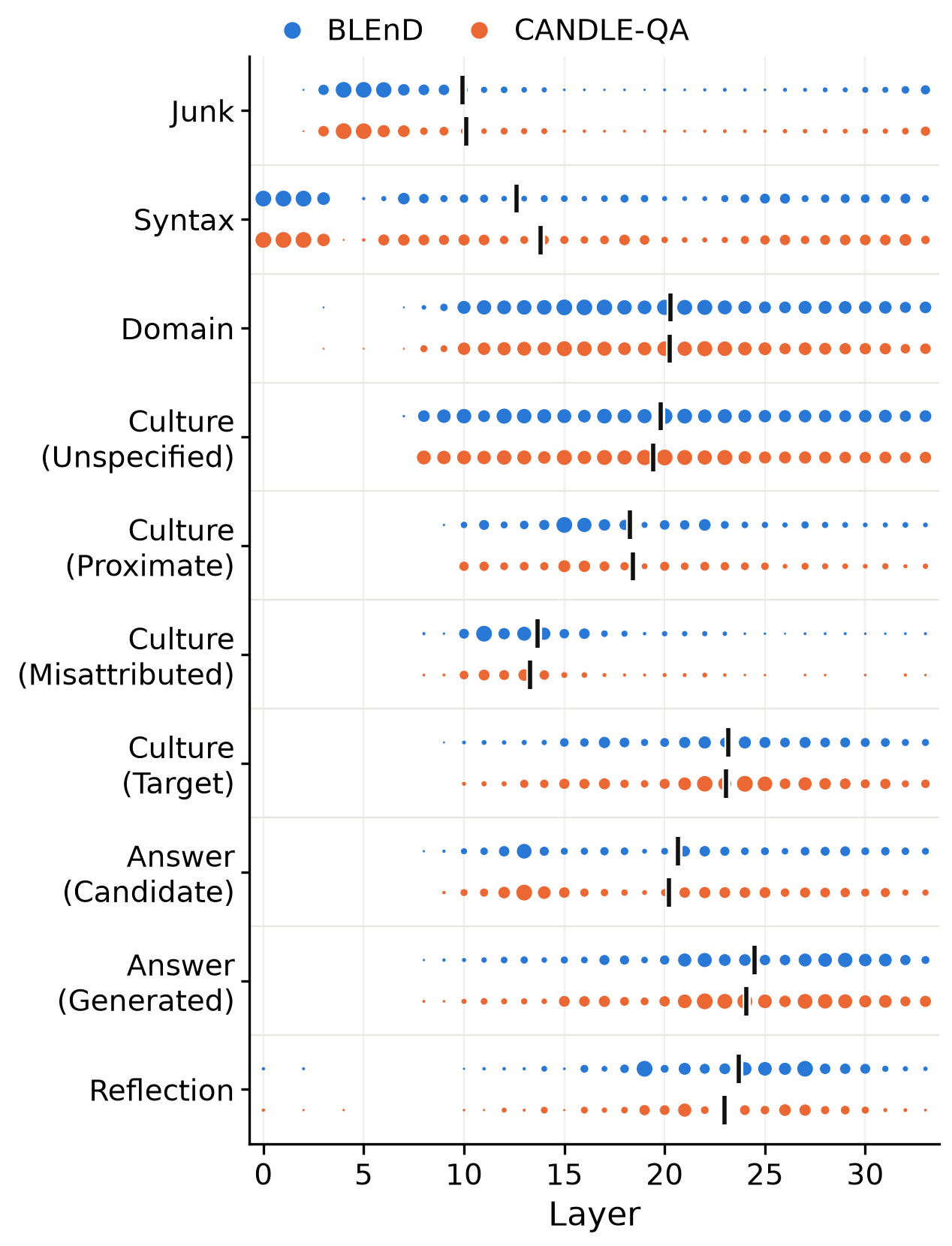}
         \caption{Result with \texttt{gemma-4b}}
         \label{fig:layer_tax_gemma}
     \end{subfigure}
     \hfill
     \begin{subfigure}[b]{0.47\textwidth}
         \centering
         \includegraphics[width=\columnwidth]{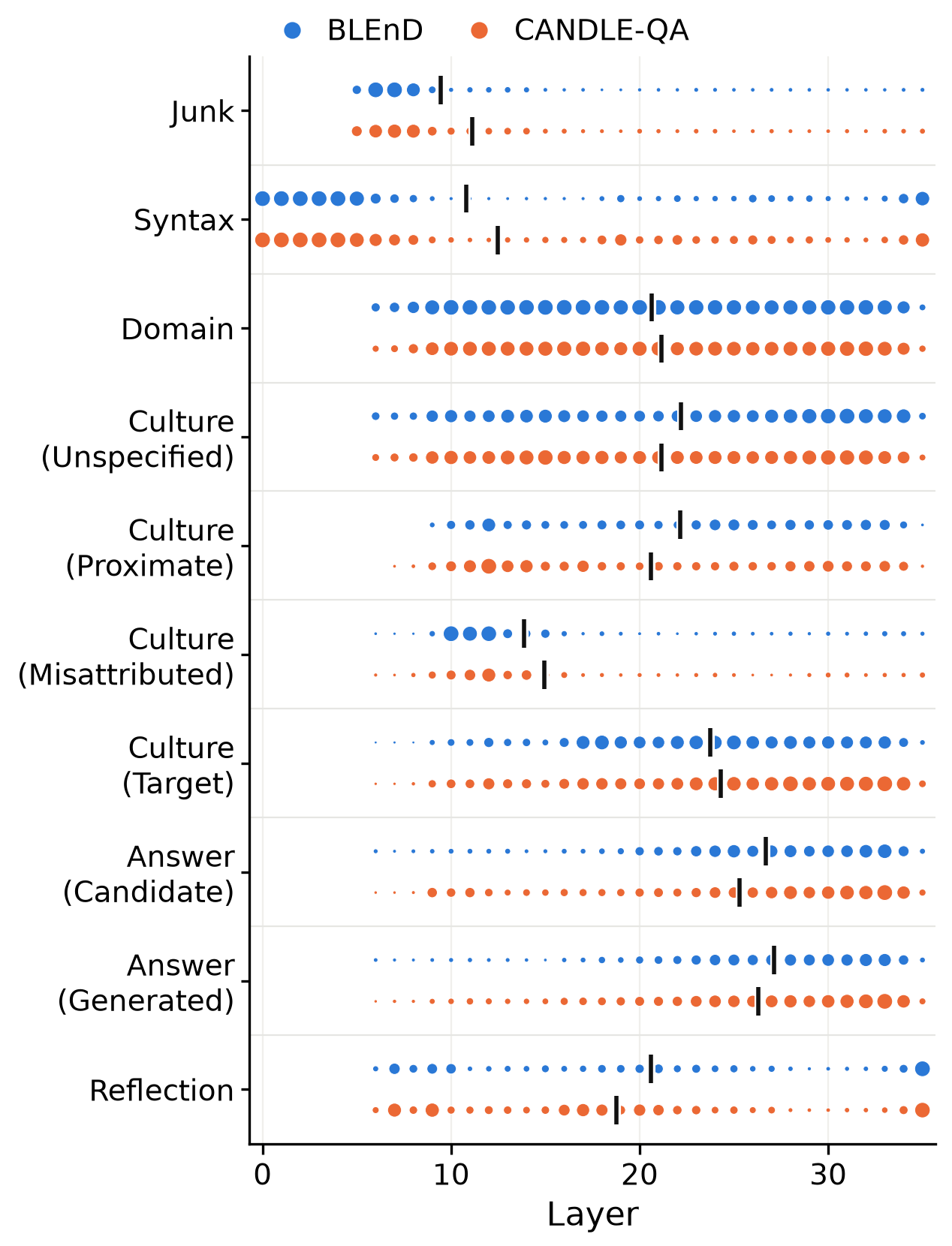}
         \caption{Result with \texttt{Ministral-8b}}
         \label{fig:layer_tax_ministral}
     \end{subfigure}
        \caption{We show the layer position at which each dimension is annotated with \texttt{gemma-4b} and \texttt{Ministral-8b}. The size of the circle scales with the number of questions that is annotated for that label. The pipe symbol represents the average layer index.}
        \label{fig:layer_tax_two}
\end{figure*}

\begin{table*}[]
\centering
\resizebox{0.75\textwidth}{!}{%
\begin{tabular}{@{}llll|llll@{}}
\toprule
\multicolumn{4}{c|}{\texttt{Ministral-8b}}                          & \multicolumn{4}{c}{\texttt{gemma-4b}}                              \\ \midrule
\multicolumn{2}{c}{BLEnD} & \multicolumn{2}{c|}{CANDLE QA} & \multicolumn{2}{c}{BLEnD} & \multicolumn{2}{c}{CANDLE QA} \\ \midrule
China        & 13.65   & China          & 14.45      & US           & 14.36   & US             & 16.61     \\
US           & 13.82   & US             & 14.69      & China        & 17.75   & Greece         & 17.01    \\
Spain        & 18.71   & Spain          & 19.13      & Spain        & 18.44   & China          & 17.90     \\
S. Korea     & 19.38     & Brazil         & 19.29      & S. Korea     & 19.31   & Spain          & 19.08     \\
Greece       & 20.60    & Greece         & 19.62      & Indonesia    & 20.44   & S. Korea       & 19.15     \\
Iran         & 21.39   & S. Korea       & 20.70      & Greece       & 21.10       & Indonesia      & 19.84     \\
Ethiopia     & 21.41   & Nepal          & 21.40      & Ethiopia     & 22.67   & Brazil         & 20.03     \\
Indonesia    & 21.61    & Indonesia      & 23.14      & Iran         & 23.56    & Iran           & 21.15     \\
Azerbaijan   & 22.62   & Iran           & 23.88      & Azerbaijan   & 24.28   & Peru           & 21.98     \\
Algeria      & 26.15   & Peru           & 24.13      & Algeria      & 25.89   & Nepal          & 22.18     \\ \bottomrule
\end{tabular}%
}
\caption{We present the mean first layer that Culture (Target) appears within intermediate cultural reasoning across questions with \texttt{gemma-4b} and \texttt{Ministral-8b}.}
\label{tab:layer_culture_reasoning}
\end{table*}

\begin{table}[]
\resizebox{\columnwidth}{!}{%
\begin{tabular}{@{}ll@{}}
\toprule
\multicolumn{1}{l|}{\textbf{Target culture}} & \textbf{Incorrectly predicted cultures (Top-5 error)} \\ \midrule
\multicolumn{2}{c}{\textbf{gemma-4b}}                                                            \\ \midrule
\multicolumn{1}{l|}{Algeria}                 & US, Middle East, Maasai, Japan, Morocco           \\
\multicolumn{1}{l|}{Azerbaijan}              & US, Maasai, Japan, Russia, Middle East            \\
\multicolumn{1}{l|}{China}                   & Japan, US, Maasai, Philippines, Ancient Egypt     \\
\multicolumn{1}{l|}{Ethiopia}                & US, Maasai, Philippines, Japan, Middle East       \\
\multicolumn{1}{l|}{Greece}                  & US, Maasai, Japan, Italy, Philippines             \\
\multicolumn{1}{l|}{Indonesia}               & US, Japan, Maasai, Philippines, Thailand          \\
\multicolumn{1}{l|}{Iran}                    & US, Japan, Middle East, Maasai, Philippines       \\
\multicolumn{1}{l|}{South Korea}             & Japan, US, Maasai, Philippines, Ancient Egypt     \\
\multicolumn{1}{l|}{Spain}                   & US, Japan, Maasai, UK, Philippines                \\
\multicolumn{1}{l|}{US}                      & Maasai, Japan, UK, Ancient Egypt, Mexico          \\ \midrule
\multicolumn{2}{c}{\textbf{Ministral-8b}}                                                        \\ \midrule
\multicolumn{1}{l|}{Algeria}                 & Inuit, Japan, France, North Africa, Morocco       \\
\multicolumn{1}{l|}{Azerbaijan}              & Inuit, Japan, Russia, Iran, US                    \\
\multicolumn{1}{l|}{China}                   & Inuit, Japan, US, Australia, Ancient Greece       \\
\multicolumn{1}{l|}{Ethiopia}                & Inuit, Japan, India, Maasai, US                   \\
\multicolumn{1}{l|}{Greece}                  & Inuit, Japan, Ancient Greece, US, France          \\
\multicolumn{1}{l|}{Indonesia}               & Inuit, Japan, China, Indian, US                   \\
\multicolumn{1}{l|}{Iran}                    & Inuit, Japan, India, US, China                    \\
\multicolumn{1}{l|}{South Korea}             & Japan, Inuit, China, US, India                    \\
\multicolumn{1}{l|}{Spain}                   & Inuit, Japan, France, US, Mexico                  \\
\multicolumn{1}{l|}{US}                      & Inuit, Japan, China, Western, Spain               \\ \bottomrule
\end{tabular}%
}
\caption{Top-5 incorrectly predicted cultures for each target culture. The result is on the BLEnD dataset.}
\label{tab:blend_error_appx}
\end{table}

\section{Results on Intermediate Cultural Reasoning}

Figure \ref{fig:layer_tax_two} shows results with \texttt{gemma-4b} and \texttt{Ministral-8b}. We see a similar order of layer reasoning in \texttt{Llama-8b}.

\begin{table}[]
\centering
\resizebox{0.8\columnwidth}{!}{%
\begin{tabular}{@{}lll@{}}
\toprule
\multicolumn{1}{c}{} & \multicolumn{1}{c}{BLEnD} & \multicolumn{1}{c}{CANDLE QA} \\ \midrule
\texttt{Llama-8b}             & 0,906               & 0,889                   \\
\texttt{gemma-4b}             & 0,758              & 0,749                  \\
\texttt{Ministral-8b}         & 0,776               & 0,754                    \\ \bottomrule
\end{tabular}%
}
\caption{We present the parsing success rate.}
\label{tab:parse_rate}
\end{table}

\begin{table}[]
\centering
\resizebox{0.9\columnwidth}{!}{%
\begin{tabular}{@{}l|lll@{}}
\toprule
\textbf{}     & \textbf{Llama-8b} & \textbf{Ministral-8b} & \textbf{gemma-4b} \\ \midrule
US & 0,583             & 0,632                 & 0,369             \\
Spain         & 0,561             & 0,432                 & 0,514             \\
South Korea   & 0,490             & 0,334                 & 0,331             \\
Brazil        & 0,469             & 0,396                 & 0,394             \\
China         & 0,461             & 0,409                 & 0,617             \\
Indonesia     & 0,423             & 0,344                 & 0,368             \\
Greece        & 0,354             & 0,317                 & 0,357             \\
Iran          & 0,340             & 0,275                 & 0,366             \\
Peru          & 0,330             & 0,197                 & 0,194             \\
Nepal         & 0,292             & 0,153                 & 0,255             \\ \bottomrule
\end{tabular}%
}
\caption{We present averaged accuracies for ten cultures with CANDLE QA.}
\label{tab:candle_acc}
\end{table}

\begin{table}[]
\resizebox{\columnwidth}{!}{%
\begin{tabular}{@{}ll@{}}
\toprule
\multicolumn{1}{l|}{\textbf{Target culture}} & \textbf{Incorrectly predicted cultures (Top-5 error)}   \\ \midrule
\multicolumn{2}{c}{\texttt{Llama-8b}}                                                              \\ \midrule
\multicolumn{1}{l|}{Brazil}                  & Japan, Spain, Mexico, Africa, Inuit                 \\
\multicolumn{1}{l|}{China}                   & Japan, Asia, Europe, Ancient Greece, Inuit          \\
\multicolumn{1}{l|}{Greece}                  & Inuit, Ancient Greece, Italy, US, Spain             \\
\multicolumn{1}{l|}{Indonesia}               & Japan, Asia, Malaysia, Australia, India             \\
\multicolumn{1}{l|}{Iran}                    & Japan, Middle East, India, Jewish, Ancient Greece   \\
\multicolumn{1}{l|}{Nepal}                   & Japan, India, Tibet, Inuit, US                      \\
\multicolumn{1}{l|}{Peru}                    & Japan, Mexico, Spain, US, Brazil                    \\
\multicolumn{1}{l|}{South Korea}             & Japan, Asia, China, Jewish, Tibet                   \\
\multicolumn{1}{l|}{Spain}                   & Japan, Italy, Basque, Mexico, France                \\
\multicolumn{1}{l|}{US}                      & Japan, UK, Africa, Western, South Korea             \\ \midrule
\multicolumn{2}{c}{\texttt{gemma-4b}}                                                              \\ \midrule
\multicolumn{1}{l|}{Brazil}                  & US, Maasai, Philippines, Japan, Mexico              \\
\multicolumn{1}{l|}{China}                   & Japan, US, Maasai, Ancient Egypt, Philippines       \\
\multicolumn{1}{l|}{Greece}                  & US, Maasai, Ancient Greece, Mediterranean, Italy    \\
\multicolumn{1}{l|}{Indonesia}               & US, Philippines, Maasai, Japan, Asia                \\
\multicolumn{1}{l|}{Iran}                    & US, Middle East, Maasai, Japan, Philippines         \\
\multicolumn{1}{l|}{Nepal}                   & US, Maasai, India, Philippines, Japan               \\
\multicolumn{1}{l|}{Peru}                    & US, Maasai, Mexico, Philippines, Japan              \\
\multicolumn{1}{l|}{South Korea}             & Japan, US, Maasai, Philippines, Ancient Egypt       \\
\multicolumn{1}{l|}{Spain}                   & US, Maasai, Italy, Japan, Mexico                    \\
\multicolumn{1}{l|}{US}                      & Maasai, UK, Japan, Ancient Egypt, Italy             \\ \midrule
\multicolumn{2}{c}{\texttt{Ministral-8b}}                                                          \\ \midrule
\multicolumn{1}{l|}{Brazil}                  & Inuit, Japan, Mexico, Spain, France                 \\
\multicolumn{1}{l|}{China}                   & Inuit, Japan, Ancient Egypt, India, Ancient Rome   \\
\multicolumn{1}{l|}{Greece}                  & Inuit, Ancient Greece, Japan, Ancient Egypt, France \\
\multicolumn{1}{l|}{Indonesia}               & Inuit, Japan, India, China, Mexico                  \\
\multicolumn{1}{l|}{Iran}                    & Inuit, Japan, India, Ancient Egypt, Middle East     \\
\multicolumn{1}{l|}{Nepal}                   & Inuit, India, Japan, China, Ancient Egypt           \\
\multicolumn{1}{l|}{Peru}                    & Inuit, Japan, Mexico, Spain, Ancient Egypt          \\
\multicolumn{1}{l|}{South Korea}             & Japan, Inuit, China, Mexico, Russia                 \\
\multicolumn{1}{l|}{Spain}                   & Inuit, Japan, Mexico, France, Netherlands           \\
\multicolumn{1}{l|}{US}                      & Inuit, Japan, France, UK, Africa                    \\ \bottomrule
\end{tabular}%
}
\caption{Top-5 incorrectly predicted cultures for each target culture. The result is on CANDLE QA. We see common incorrectly predicted cultures across models: Japan for \texttt{Llama-8b}, US for \texttt{gemma-4b}, and Inuit for \texttt{Ministral-8b}.}
\label{tab:candle_error}
\end{table}

\section{Target Culture Identification}

\subsection{Parsing Implementation Details}\label{appx:parse}

Figure \ref{fig:parse_dictionary} presents the dictionary we use for parsing culture names from CulTrace generations. Table \ref{tab:parse_rate} shows the parsing success rate on each model and dataset with the dictionary.

\subsection{Results}\label{appx:culture_acc}

Table \ref{tab:candle_acc} shows averaged accuracies with CANDLE QA. We see the US as the highest on \texttt{Llama-8b} and \texttt{Ministral-8b}. China achieves highest with \texttt{gemma-4b}. Table \ref{tab:blend_error_appx} and Table \ref{tab:candle_error} show top-5 incorrectly predicted cultures.

\begin{table*}[t]
\centering
\small
\begin{tabularx}{\textwidth}{@{}lX@{}}
\toprule
Question   & Based on the characteristic population curves that result from plotting population growth of a species, the most effective means of controlling the mosquito population is to \\ \midrule
LLM Answer & D. increase the mortality rate. Reasoning: Population growth curves, particularly logistic curves, illustrate the interplay between population size and environmental carrying capacity (K). \\ \midrule
Layer 0    & The assistant provides the most suitable option by analysing the context and selecting the response that best fits the situation. \\
Layer 1    & The assistant provides the most appropriate option from the given choices, along with a detailed explanation of its reasoning. \\
Layer 2    & The assistant provides the most suitable option by evaluating each choice and selecting the one that best fits the context. \\
Layer 3    & The assistant provides the most suitable option from the given choices, along with a detailed explanation of its reasoning. \\
Layer 4    & The assistant provides the most appropriate option by evaluating each choice and selecting the one that best aligns with the user's needs. \\
Layer 5    & The assistant provides the most suitable option from the given choices, explaining its reasoning to justify the selection. \\
Layer 6    & The assistant provides the most appropriate option as the answer, explaining its reasoning to justify the choice. \\
Layer 7    & The assistant provides the most suitable option by evaluating the context and selecting the response that best fits the situation. \\
Layer 8    & The assistant provides the most suitable option by evaluating the context and selecting the one that best fits the scenario. \\
Layer 9    & The assistant provides the most appropriate option by evaluating the context and selecting the answer that best fits the scenario. \\
Layer 10   & The assistant provides the most suitable option by evaluating each choice and selecting the one that best addresses the user's query. \\
Layer 11   & The assistant provides the most appropriate option by analysing the context and selecting the answer that best fits the scenario. \\
Layer 12   & The assistant provides the most appropriate option, which is the one that directly addresses the user's query about the relationship between the variables. \\
Layer 13   & The assistant provides the answer by selecting the option that best aligns with the principles of thermodynamics, specifically focusing on the conservation of energy. \\
Layer 14   & The assistant provides the most appropriate option by evaluating each choice against the context of the question. \\
Layer 15   & The assistant provides the most appropriate answer by explaining the reasoning behind its choice, ensuring the user understands the rationale. \\
Layer 16   & The assistant provides the answer by explaining the reasoning behind choosing the most effective strategy to reduce the population. \\ \bottomrule
\end{tabularx}
\caption{An example of CulTrace outputs with patching representations of an MMLU science question from \texttt{gemma-4b} (layers 0--16). Layers 17--33 are shown in Table~\ref{tab:mmlu_gemma_example_b}.}
\label{tab:mmlu_gemma_example_a}
\end{table*}

\begin{table*}[t]
\centering
\small
\begin{tabularx}{\textwidth}{@{}lX@{}}
\toprule
Question   & Based on the characteristic population curves that result from plotting population growth of a species, the most effective means of controlling the mosquito population is to \\ \midrule
LLM Answer & D. increase the mortality rate. Reasoning: Population growth curves, particularly logistic curves, illustrate the interplay between population size and environmental carrying capacity (K). \\ \midrule
Layer 17   & The assistant should provide the answer by explaining the reasoning behind choosing the most effective method to control the population size. \\
Layer 18   & The assistant provides the answer by explaining the reasoning behind it, ensuring the user understands the rationale. \\
Layer 19   & The assistant provides the answer by explaining the reasoning behind its choice, ensuring the user understands the rationale. \\
Layer 20   & The assistant provides the answer by explaining the reasoning behind the most effective method for controlling population growth. \\
Layer 21   & The assistant provides the answer by explaining the reasoning behind it, ensuring the user understands the logic. \\
Layer 22   & The assistant provides the most appropriate answer by explaining the reasoning behind it, ensuring the user understands the rationale. \\
Layer 23   & The assistant provides the most appropriate answer by explaining the reasoning behind it, ensuring the user understands the rationale. \\
Layer 24   & The assistant provides the most appropriate answer by explaining the reasoning behind it, ensuring the user understands the rationale. \\
Layer 25   & The assistant provides the answer by explaining the reasoning behind the most effective population control strategy. \\
Layer 26   & The assistant provides the most appropriate answer by explaining the reasoning behind it, ensuring the user understands the logic. \\
Layer 27   & The assistant provides the answer by explaining the reasoning behind the most effective control method for population growth. \\
Layer 28   & The assistant provides the answer that the most effective way to control population growth is to limit the number of offspring each individual can have. \\
Layer 29   & The assistant provides the answer `Population control' as the most appropriate solution, explaining that it is the most effective way to manage the population of invasive species. \\
Layer 30   & The assistant provides the most appropriate answer by explaining the reasoning behind it, ensuring the user understands the rationale. \\
Layer 31   & The answer is that the most appropriate option is to reduce the population of invasive species to prevent overpopulation and ecological imbalance. \\
Layer 32   & The answer is that the most appropriate response is `None', because the population growth is exponential and cannot be sustained indefinitely. \\
Layer 33   & The assistant provides the answer that the most appropriate solution is to allow the population to grow without constraints, as this is the most logical approach. \\ \bottomrule
\end{tabularx}
\caption{Continuation of Table~\ref{tab:mmlu_gemma_example_a}: CulTrace outputs for layers 17--33.}
\label{tab:mmlu_gemma_example_b}
\end{table*}

\begin{figure*}[h!]
    \centering
    \begin{sharp_box}\fontsize{9pt}{11.5pt}\selectfont

    {
    ``american": ``United States", ``british": ``United Kingdom",
    ``french": ``France", ``spanish": ``Spain", ``italian": ``Italy", ``dutch": ``Netherlands",
    ``irish": ``Ireland", ``scottish": ``Scotland", ``welsh": ``Wales", ``greek": ``Greece",
    ``german": ``Germany", ``norwegian": ``Norway", ``swedish": ``Sweden", ``danish": ``Denmark",
    ``portuguese": ``Portugal", ``swiss": ``Switzerland", ``czech": ``Czech Republic",
    ``russian": ``Russia", ``serbian": ``Serbia", ``albanian": ``Albania", ``georgian": ``Georgia",
    ``turkish": ``Turkey", ``azerbaijani": ``Azerbaijan", ``kazakh": ``Kazakhstan",
    ``turkmen": ``Turkmenistan", ``mongolian": ``Mongolia", ``chinese": ``China",
    ``japanese": ``Japan", ``korean": ``South Korea", ``indian": ``India", ``tibetan": ``Tibet",
    ``filipino": ``Philippines", ``ethiopian": ``Ethiopia", ``algerian": ``Algeria",
    ``moroccan": ``Morocco", ``egyptian": ``Egypt", ``sudanese": ``Sudan", ``somali": ``Somalia",
    ``ghanaian": ``Ghana", ``malian": ``Mali", ``south african": ``South Africa",
    ``mexican": ``Mexico", ``argentinian": ``Argentina", ``australian": ``Australia",
    ``indonesian": ``Indonesia", ``iranian": ``Iran", ``persian": ``Iran",
    ``kenyan": ``Kenya", ``brazilian": ``Brazil", ``armenian": ``Armenia",
    ``nepalese": ``Nepal", ``nepali": ``Nepal", ``peruvian": ``Peru",
    ``andean": ``Peru",
    ``nigerian": ``Nigeria", ``romanian": ``Romania", ``ukrainian": ``Ukraine",
    ``austrian": ``Austria", ``malaysian": ``Malaysia", ``pakistani": ``Pakistan",
    ``afghan": ``Afghanistan", ``macedonian": ``North Macedonia",
    ``slovenian": ``Slovenia", ``hungarian": ``Hungary", ``canadian": ``Canada",
    ``thai": ``Thailand",
    ``african": ``Africa", ``north african": ``North Africa", ``european": ``Europe",
    ``eastern european": ``Eastern Europe", ``asian": ``Asia", ``middle eastern": ``Middle East",
    ``scandinavian": ``Scandinavia", ``mediterranean": ``Mediterranean",
    }
    \end{sharp_box}
    \caption{We present the dictionary used to parse CulTrace generations for target culture identification.}
    \label{fig:parse_dictionary}
\end{figure*}

\begin{figure*}[h!]
    \centering
    \begin{sharp_box}\fontsize{9pt}{11.5pt}\selectfont

        ``Do not use any markdown or special formatting. Question: \{Question\}\textbackslash nThink about the answer, provide it in a single term, followed by your reasoning.''
    \end{sharp_box}
    \caption{An input prompt used during the Representation extraction stage for the target LLM.}
    \label{fig:appendix_Prompts}
\end{figure*}

\begin{figure*}[h!]
    \centering
    \begin{sharp_box}\fontsize{9pt}{11.5pt}\selectfont

    You rewrite fill-in-the-blank sentences about cultural knowledge into natural questions.

You are given a country, a topic, a gold answer, and a cloze sentence in which the gold answer has been replaced by the marker [blank]. Write one question whose correct answer is exactly the gold answer.

Rules:
1. The question must be answerable with the gold answer, and the gold answer must be the single most natural answer to it.
2. Keep every descriptive cue that the sentence provides, including the country, culture, or region it names. The question must be no easier and no harder than the cloze: it carries the same information, only re-phrased.
3. Never write the gold answer, a translation of it, an inflected form of it, or a near-synonym that gives it away, anywhere in the question.
4. Add no facts that are not in the sentence, and drop none that are.
5. Output exactly one question: a single sentence, ending in a question mark, no quotes, no preamble, no explanation.
6. Do not mention [blank], the sentence, or the task itself.
7. When the gold answer is the name of a thing, prefer the natural ``What is ... called?'' / ``What is the name of ...?'' phrasing over an awkward inversion.
8. Keep the tense and number of the original sentence (a sentence about the past asks a past-tense question).
9. If the sentence contains [blank] more than once, every occurrence stands for the same gold answer. Still write only one question.

Answer with a JSON object of the form {``question'': ``<the question>''}.

    \end{sharp_box}
    \caption{An input prompt used during the Representation extraction stage for the target LLM.}
    \label{fig:convert_prompt}
\end{figure*}

\begin{figure*}[h!]
    \centering
    \begin{sharp_box}\fontsize{9pt}{11.5pt}\selectfont

    ``""You are an expert annotator for a research project studying how language models process cultural knowledge internally. You will receive a single layer description — a natural language verbalization of what an LLM is representing at a specific layer — along with metadata about the question being answered. Your task is to annotate the layer across six categories.

\#\# Category Definitions

\#\#\# Culture Processing

Captures whether the layer is engaging with cultural information, and if so, how accurately it identifies the target culture.

- `Target`: The correct target culture is explicitly named or unambiguously referenced (e.g., ``Iranian society `` when the country is Iran, ``Ethiopian cuisine `` when the country is Ethiopia).

- `Proximate`: A culture that is geographically, linguistically, or typologically related to the target is named (e.g., ``Middle Eastern traditions `` for Iran, ``Japanese culture `` for South Korea, ``East African customs `` for Ethiopia). The named culture must share a plausible regional or cultural-historical connection with the target.

- `Misattributed`: A specific culture is named that has no meaningful proximity to the target (e.g., ``Japanese society `` for Ethiopia, ``Scottish traditions `` for Indonesia, ``Japanese art of origami `` for Iran).

- `Unspecified`: The layer uses cultural language — words like ``cultural significance, `` ``traditions, `` ``heritage, `` ``cultural identity, `` ``cultural context `` — but does not name or imply any specific culture or region.

- Empty: The layer contains no cultural processing at all. Note: merely discussing a topic (e.g., food, holidays) without cultural framing is Domain, not Culture: Unspecified.

\#\#\# Domain Processing

Captures whether the layer engages with the topic area of the question.

- `x`: The layer engages with either the question's core subject matter or the broader topic area surrounding it. For example, for a question about a sport commentator, both ``a particular commentator's style `` and ``a player's decision to retire from soccer `` count as domain processing.
- Empty: No engagement with the topic area.

\#\#\# Answer Processing

Captures whether a specific answer to the question is identifiable in the layer.

- `Generated`: The model's actual output answer is specifically identifiable — either named explicitly or described with enough specificity to uniquely identify it. Be conservative: vague references to the topic do not count. ``A traditional Ethiopian dish `` does NOT qualify as Generated even if the model answered ``Injera. `` ``Traditional Ethiopian foods, particularly injera `` DOES qualify. The test is: could you identify the specific answer from the description alone?

- `Candidate`: A plausible alternative answer to the question surfaces — a specific, identifiable item distinct from the model's generated answer. The same test as Generated applies: could you identify the specific alternative from the description alone? A named entity qualifies (e.g., ``moussaka `` when the model answered ``Gyro"). The named entity does not need to be framed as an answer — a layer simply discussing a specific instance of the question's answer type counts (e.g., a layer about ``the sport of cycling `` is a Candidate when the question asks which sport and the model answered ``Water Polo"). A superlative or descriptive phrase that does not uniquely identify an entity — ``the most successful football club, `` ``a famous local team, `` ``the most iconic superhero `` — does NOT qualify; that is Domain processing. In particular, a phrase that merely restates or generalizes the question's own framing (e.g., ``the most successful club in the world `` for a question about the most popular team in a country) is not an answer at all. The test is naming, not framing: named specific entity of the answer type → Candidate; unnamed description → Domain at most.

- Empty: No specific answer is identifiable.
 ``""
    
    \end{sharp_box}
    \caption{An input prompt used for annotating intermediate cultural reasonings.}
    \label{fig:judge_prompt_1}
\end{figure*}

\begin{figure*}[h!]
    \centering
    \begin{sharp_box}\fontsize{9pt}{11.5pt}\selectfont

\#\#\# Reflection

Captures meta-cognitive processing — the model reasoning about the *substance* of its own answer or reasoning.

- `x`: The layer describes the model evaluating, justifying, or revisiting the reasoning behind a specific response it has given or is about to give. The layer must take its own reasoning or answer as the object of thought. Indicators include: ``the reasoning behind its previous response, `` ``ensuring the answer is logically sound, `` ``logical justification for its conclusion, `` ``weighing whether its answer is correct."

- Empty: everything else, including the two cases below.

Reflection is NOT: generic statements about how the response should be phrased (→ Syntax: `Task/Response`), nor restatements of the prompt or the annotation situation (→ Syntax: `Echo`). Both describe the response or the task rather than evaluating any particular reasoning; their indicators are listed under Syntax Processing below.

The test: could you name the specific reasoning or answer being evaluated? If not, it is not Reflection.

\#\#\# Syntax Processing

Captures layers about *how to respond* rather than *what the answer is* — the task format, the response as an object, or the prompt framing. Use the sub-values below so that downstream analysis can separate genuine task processing from degenerate self-description.

- `x`: representing what the task requires, the shape of the expected output, or the manner in which the response should be phrased — with no reference to the answer's content. This covers both the format of the task and the style of the response; do not try to separate them. Indicators include: ``the impact of a single word in a sentence `` (for fill-in-the-blank tasks), ``the constraints of the task, `` ``selecting a single word to fill the blank, `` ``how to convey a message in a single word, followed by a brief explanation, `` ``the importance of brevity and clarity in its responses, `` ``how to convey its thoughts in a concise manner, `` ``prepared to articulate in a concise manner, `` ``how to provide a helpful and accurate response."

\#\#\# Junk
Captures layer descriptions unrelated to any aspect of the task.

- `x`: The description is coherent but has no connection to the question, answer, domain, culture, or task structure. Common patterns include abstract philosophical musings about time, infinity, perception, or consciousness.

- Empty: The layer is relevant to at least one other category.

- A layer describing the task, the prompt, or how the response should be phrased is Syntax, not Junk — however contentless it looks. Junk is for content about something else entirely.

- **Junk is mutually exclusive with all other categories.** If junk is `x`, all other fields must be empty.

\#\# Annotation Guidelines

1. Every layer must receive at least one label. If no other category applies, the layer is Junk — but check Syntax first, see guideline 8. An annotation with every field empty is invalid — never output one.
2. Multiple categories can fire simultaneously, except for junk. A layer can have culture, domain, answer, reflection, and syntax labels all at once.
3. When in doubt between a more specific and a less specific label, prefer the less specific one.
4. Proximate vs. Misattributed depends on the target country. The same named culture can be Proximate for one target and Misattributed for another (e.g., ``Japanese `` is Proximate for South Korea but Misattributed for Ethiopia).
5. Do not infer answer labels from domain processing. A layer discussing ``food `` for a food question is Domain, not Answer — unless a specific food item matching the model's answer is identifiable. This applies equally to Candidate: ``the best/most popular/most successful X `` without a name identifying which X is Domain, not Candidate.
6. Culture: Unspecified requires explicit cultural language. A layer about ``a particular dish `` is Domain only. A layer about ``the cultural significance of a particular dish `` is both Culture: Unspecified and Domain.
7. Domain is broad. Content tangentially related to the topic area still counts. Junk is reserved for content with no connection to any aspect of the task.
8. Reflection, Syntax and Junk are decided in this order. If the layer evaluates the reasoning behind a specific answer → Reflection. Otherwise, if it is about the task format, the phrasing of the response, or the prompt itself → Syntax. Only if it is about none of these → Junk. A contentless layer is far more often Syntax than Reflection.

\#\# Output Format

Respond with ONLY a JSON object. No explanation, no reasoning, no markdown formatting.
{"culture": ``", ``domain": ``", ``answer": ``", ``reflection": ``", ``syntax": ``", ``junk": ``"}

Use exact label strings: `x` for domain, reflection, and junk.
 ``""
    
    \end{sharp_box}
    \caption{An input prompt used for annotating intermediate cultural reasonings.}
    \label{fig:judge_prompt_2}
\end{figure*}

\end{document}